\documentclass[10pt]{article} 
\usepackage[accepted]{tmlr}

\usepackage{amsmath,amsfonts,bm}

\def\eqref#1{equation~\ref{#1}}

\def\1{\bm{1}}

\DeclareMathAlphabet{\mathsfit}{\encodingdefault}{\sfdefault}{m}{sl}
\SetMathAlphabet{\mathsfit}{bold}{\encodingdefault}{\sfdefault}{bx}{n}

\usepackage{url}
\usepackage{booktabs}       
\usepackage{amsfonts}       
\usepackage{nicefrac}       
\usepackage{microtype}      
\usepackage{xcolor}         
\usepackage[colorlinks=true, linkcolor=blue, urlcolor=magenta, citecolor=blue]{hyperref}

\usepackage{algorithm}
\usepackage{algcompatible}

\usepackage{bookmark}
\usepackage{amsmath}
\usepackage{amsthm}
\usepackage{amssymb}
\usepackage{graphicx}
\usepackage{multirow}
\usepackage{wrapfig}
\usepackage{floatflt}

\definecolor{mine}{RGB}{205, 232, 248}%

\newcommand{\scr}[1]{{\scriptsize #1}}

\title{Decentralized Transformers with Centralized Aggregation are Sample-Efficient Multi-Agent World Models}

\author{\name Yang Zhang\textsuperscript{1}\thanks{Work done during Yang Zhang's internship on both Shanghai AI Lab and TeleAI.}~, Chenjia Bai\textsuperscript{2}\thanks{Corresponding Authors.}~, Bin Zhao\textsuperscript{3}, Junchi Yan\textsuperscript{4}, Xiu Li\textsuperscript{1\dag}, Xuelong Li\textsuperscript{2} \\
\addr \textsuperscript{1}Tsinghua University~~~\textsuperscript{2}Institute of Artificial Intelligence (TeleAI), China Telecom \\
\textsuperscript{3}Shanghai AI Laboratory~~~\textsuperscript{4}Shanghai Jiaotong University}

\def\month{04}  
\def\year{2025} 
\def\openreview{\url{https://openreview.net/forum?id=xT8BEgXmVc}} 

\begin{document}

\maketitle

\begin{abstract}
Learning a world model for model-free Reinforcement Learning (RL) agents can significantly improve the sample efficiency by learning policies in imagination. However, building a world model for Multi-Agent RL (MARL) can be particularly challenging due to the scalability issue across {different number of agents} in a centralized architecture, and also the non-stationarity issue in a decentralized architecture stemming from the inter-dependency among agents. To address both challenges, we propose a novel world model for MARL that learns decentralized local dynamics for scalability, combined with a centralized representation aggregation from all agents. We cast the dynamics learning as an auto-regressive sequence modeling problem over discrete tokens by leveraging the expressive Transformer architecture, in order to model complex local dynamics across different agents and provide accurate and consistent long-term imaginations. As the first pioneering Transformer-based world model for multi-agent systems, we introduce a Perceiver Transformer as an effective solution to enable centralized representation aggregation within this context. {Extensive results on Starcraft Multi-Agent Challenge (SMAC) and MAMujoco demonstrate superior sample efficiency and overall performance compared to strong model-free approaches and existing model-based methods}. Our code is available at \url{https://github.com/breez3young/MARIE}.
\end{abstract}

\section{Introduction}
Multi-Agent Reinforcement Learning (MARL) has made remarkable progress, driven largely by model-free algorithms \citep{nguyen20marl_review}.
However, due to the complexity of multi-agent systems arising from partial observability and {multiple agents' inter-dependencies that induce rapidly growing state-action space complexities}, such algorithms usually demand extensive interactions to learn coordinative behaviors \citep{Hernandez20masurvey}.
A promising solution is building a world model that approximates the environment, which has exhibited its superior sample efficiency compared to model-free approaches in single-agent RL \citep{hafner2019dreamer, Kaiser2020SimPLe, hafner2020dreamerv2, hafner2023dreamerv3, Hansen2022tdmpc, hansen2024tdmpc2}.
However, extending the design of world model in the single-agent domain to the multi-agent context encounters {unique} challenges due to the {inter-agent interactions} inherent to multi-agent environments.

The challenges primarily stem from two different means for multi-agent dynamics learning: \emph{centralized} and \emph{decentralized}.
{\emph{Centralized} world models can naturally encapsulate agent inter-dependency through joint dynamics modeling, but suffer from the combinatorial explosion of the joint action space. This leads to exponential growth in sample complexity as the number of agents increases \citep{Hernandez20masurvey, nguyen20marl_review}, making them impractical for robustly scaling across diverse team configurations (see Figure~\ref{fig:central_ablation_result} in our ablation studies).}
Conversely, applying a \emph{decentralized} world model to approximating the local dynamics of each agent mitigates the scalability issue yet incurs non-stationarity, as unexpected interventions from other agents may occur in each agent's individual environment \citep{Oliehoek16dec_pomdp}.
Furthermore, beyond these unique challenges inherent in modeling multi-agent dynamics, existing model-based MARL approaches \citep{Willemsen21mambpo, Egorov22mamba, xu2022mbvd} excessively neglect the fact that the policy learned in imaginations of the world model heavily relies on the quality of imagined trajectories \citep{alonso2023iris}.
It thereby necessitates accurate long-term predictions, especially with respect to the non-stationary local dynamics.

Inspired by the capability of Transformer~\citep{vaswani17transformer} in modeling complex discrete sequences and long-term dependency \citep{brown20llmfewshot, devlin19bert, alonso2023iris}, we seek to construct a Transformer-based world model within the multi-agent context for \emph{decentralized} local dynamics together with \emph{centralized} feature aggregation, combining the benefits of two distinct designs.

In this paper, we introduce MARIE (Multi-Agent auto-Regressive Imagination for Efficient learning), the first Transformer-based multi-agent world model for sample-efficient policy learning. Specifically, \textbf{the highlights of this paper are:}
\begin{enumerate}
    \item To tackle the inherent challenges within the multi-agent context, we build an effective world model via scalable \emph{decentralized} dynamics modeling and essential \emph{centralized} representation aggregating, which mirrors the principle of Centralized Training and Decentralized Execution (CTDE). 
    \item To enable accurate and consistent long-term imaginations from the non-stationary local dynamics, we cast the \emph{decentralized} dynamics learning as sequence modeling over discrete tokens by leveraging highly expressive Transformer architecture as the backbone. In particular, we successfully present the first Transformer-based world model for multi-agent systems. 
    \item While it remains open for how to effectively enable \emph{centralized} representation with the Transformer as the backbone, we achieve it by innovatively introducing a Perceiver Transformer \citep{jaegle2021perceiver} for efficient global information aggregation across all agents. 
    \item Experiments on the Starcraft Multi-Agent Challenge (SMAC) benchmark in low data regime and additional experiments on MAMujoco show MARIE outperforms both model-free and existing model-based MARL methods w.r.t. both sample efficiency and overall performance and demonstrate the effectiveness of MARIE. 
\end{enumerate}

\section{Related Works and Preliminaries}\label{sec:preliminary}
\textbf{Multi-Agent Reinforcement Learning.} In a model-free setting, a typical approach for cooperative MARL is centralized training and decentralized execution (CTDE), which tackles the scalability and non-stationarity issues in MARL. During the training phase, it leverages global information to facilitate agents' policy learning; while during the execution phase, it blinds itself and has only access to the partial observation around each agent for multi-agent decision-making. Model-free MARL methods with this paradigm can be divided into 2 categories: value-based \citep{Sunehag2017vdn, rashid18qmix, son19qtran, wang2021qplex} and policy-based \citep{lowe17maac, Foerster18coma, iqbal19maac, Ryu2019MultiAgentAW, liu20maga, kuba2021trust, peng2021facmac, yu2022mappo, zhang2024read, zhang2024provably}. In contrast to model-free approaches, model-based MARL algorithms remain fairly understudied. MAMBPO \citep{Willemsen21mambpo} incorporates MBPO-style \citep{janner2019mbpo} techniques into multi-agent policy learning under the CTDE framework. Tesseract \citep{mahajan21tesseract} introduces the tensorised Bellman equation and evaluates the Q-value function using Dynamic Programming (DP) together with an estimated environment model. Similar to our setting where agents learn inside of an approximate world model, MAMBA \citep{Egorov22mamba} integrates the backbone proposed in DreamerV2~\citep{hafner2020dreamerv2} with an attention mechanism across agents to sustain an effective world model in environments with an arbitrary number of agents, which leads to notably superior sample efficiency to existing model-free approaches.
In terms of model-based algorithm coupled with planning, MAZero \citep{liu2024mazero} expands the MCTS planning-based Muzero \citep{Schrittwieser2019Muzero} framework to the model-based MARL settings. 
However, learning-based or planning-based policies in these two approaches are both overly coupled with their world models, downgrading their inference efficiency and further limiting expansion in combinations with other popular model-free approaches.
In contrast, the world model in our MARIE is solely utilized to accelerate policy learning during the training phase, rather than enhancing the policy with additional informative features for decision-making. Consequently, MARIE can be seamlessly integrated with any model-free algorithm, enabling the rapid deployment and inference of a well-trained policy without requiring an accompanying world model.

Besides, a recent method CoDreamer \citep{toledo2024codreamer} extends DreamerV3 \citep{hafner2023dreamerv3} to the multi-agent setting with a graph neural network as a communication module. A related in-depth discussion is deferred to \S\ref{appendix:additional_discussion}.

\textbf{Learning behaviors within the imagination of world models.} The Dyna architecture \citep{Sutton1991Dyna} first emphasizes the utility of an estimated dynamics model in facilitating the training of the value function and policy. Inspired by the cognitive system of human beings, the concept of world model \citep{david2018worldmodels} is initially introduced by composing a variational Auto-Encoder (VAE) \citep{kingma2022vae} and a recurrent network to mimic the complete environmental dynamics, then an artificial agent is trained entirely inside the hallucinated imagination generated by the world model. SimPLe~\citep{Kaiser2020SimPLe} shows that a PPO policy~\citep{Schulman2017ppo} learned in a predictive model delivered a super-human performance in Atari domains. Dreamer~\citep{hafner2019dreamer} built the world model upon a Recurrent State Space Model (RSSM) \citep{hafner19PlaNet} that combines the deterministic latent state with the stochastic latent state to allow the model to not only capture multiple futures but also remember information over multi-steps. DreamerV2 \citep{hafner2020dreamerv2} further demonstrates the advantage of discrete latent states over Gaussian states. For MARL, MAMBA \citep{Egorov22mamba} extends DreamerV2 to multi-agent contexts by using RSSM, underscoring the potential of multi-agent learning in the imagination of world models. Recently, motivated by the success of the Transformer~\citep{vaswani17transformer}, TransDreamer \citep{chen2022transdreamer} and TWM \citep{robine2023twm} explored variants of DreamerV2, wherein the backbones of the world model were substituted with Transformers. Instead of incorporating deterministic and stochastic latent states, IRIS \citep{alonso2023iris} applies the Transformer to directly modeling sequences of observation tokens and actions of single-agent RL and achieves impressive results on Atari-100k. In contrast, the proposed MARIE concentrates on establishing effective Transformer-based world models in multi-agent contexts with shared dynamics and global representations.

\textbf{Preliminaries.} We focus on fully cooperative multi-agent systems where all agents share a team reward signal. We formulate the system as a decentralized partially observable Markov decision process (Dec-POMDP) \citep{Oliehoek16dec_pomdp}, which can be described by a tuple $(\mathcal{N}, \mathcal{S}, \boldsymbol{\mathcal{A}}, P, R, \boldsymbol{\Omega}, \boldsymbol{\mathcal{O}}, \gamma)$. $\mathcal{N} = \{1, ..., n\}$ denotes a set of agents, $\mathcal{S}$ is the finite global state space, $\boldsymbol{\mathcal{A}} = \prod_{i=1}^{n}\mathcal{A}^{i}$ is the product of finite actions spaces of all agents, i.e., the joint action space, $P : \mathcal{S} \times \boldsymbol{\mathcal{A}} \times \mathcal{S} \rightarrow [0, 1]$ is the global transition probability function, $R : \mathcal{S} \times \boldsymbol{\mathcal{A}} \rightarrow \mathbb{R}$ is the shared reward function, $\boldsymbol{\Omega} = \prod_{i=1}^{n}\Omega^{i}$ is the product of finite observation spaces of all agents, i.e., the joint observation space, $\boldsymbol{\mathcal{O}} = \{ \mathcal{O}^i, i \in \mathcal{N}\}$ is the set of observing functions of all agents. $\mathcal{O}^i : \mathcal{S} \rightarrow \Omega^i$ maps global states to the observations for agent $i$, and $\gamma$ is the discount factor. Given a global state $s_t$ at timestep $t$, agent $i$ is restricted to obtaining solely its local observation $o_{t}^{i} = \mathcal{O}^{i}(s_t)$, takes an action $a_{t}^{i}$ drawn from its policy $\pi^{i}(\cdot | o_{\leq t}^{i})$ based on the history of its local observations $o_{\leq t}^{i}$, which together with other agents' actions gives a joint action $\boldsymbol{a}_{t} = ( a_{t}^{1}, ..., a_{t}^{n} ) \in \boldsymbol{\mathcal{A}}$, equivalently drawn from a joint policy $\boldsymbol{\pi}(\cdot | \boldsymbol{o}_{\leq t}) = \prod_{i = 1}^{n} \pi^{i} (\cdot | o_{\leq t}^{i})$. Then the agents receive a shared reward $r_t = R(s_t, \boldsymbol{a}_t)$, and the environment moves to next state $s_{t+1}$ with probability $P(s_{t+1} | s_t, \boldsymbol{a}_t)$. The aim of all agents is to learn a joint policy $\boldsymbol{\pi}$ that maximizes the expected discounted return $J(\boldsymbol{\pi}) = \mathbb{E}_{s_0, \boldsymbol{a}_0, ... \sim \boldsymbol{\pi}} \left[ \sum_{t = 0}^{\infty} \gamma^t R(s_t, \boldsymbol{a}_t) \right]$.

\begin{figure*}[t!]
\vskip -0.05in
\begin{center}
\centerline{
\includegraphics[width=1.0\columnwidth]{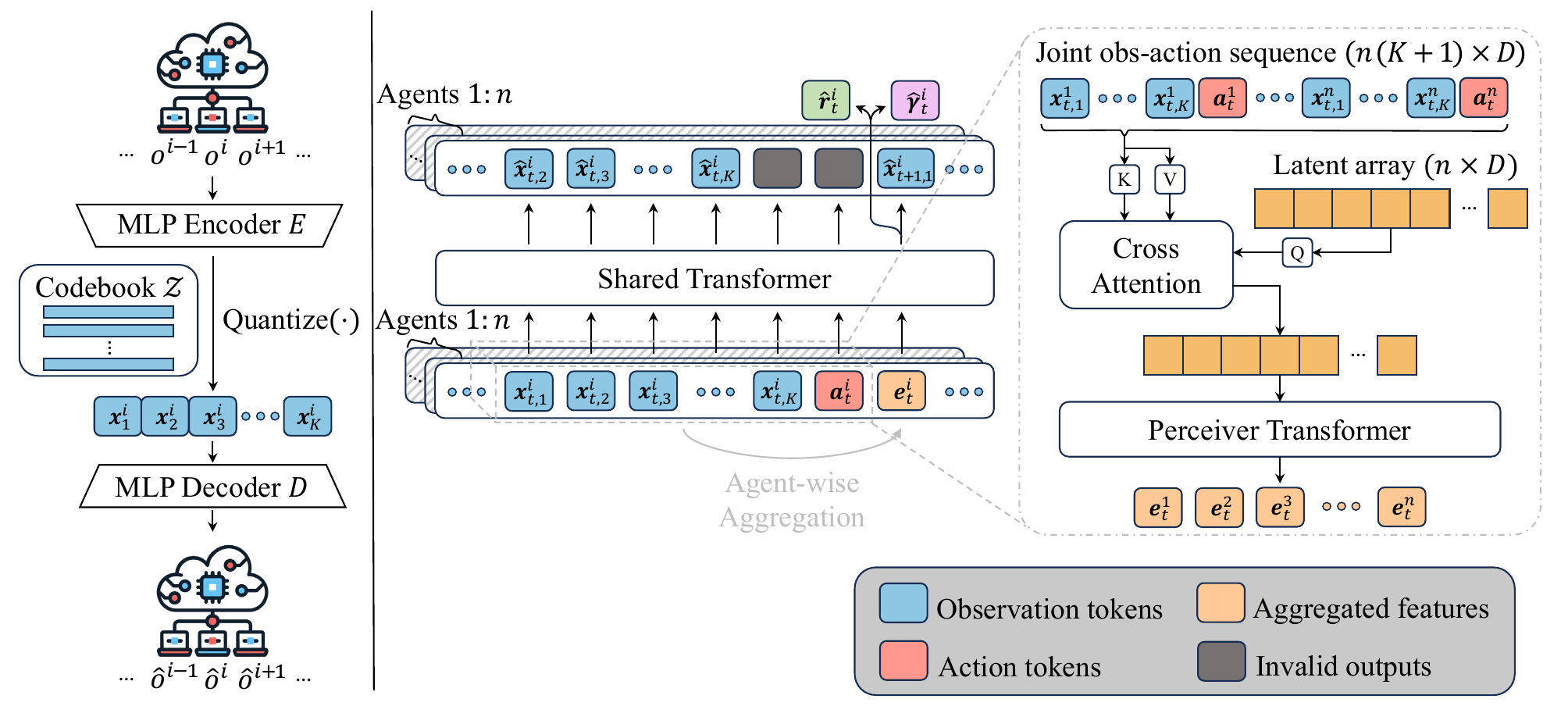}}
\vspace{-0.5em}
\caption{{\bf Overview of the proposed world model architecture in MARIE.}
VQ-VAE ({\bf \emph{left}}) maps local observations $o^{i}$ of each agent $i$ into discrete latent codes $(x^{i}_{1}, ..., x^{i}_{K})$, where $(E, D, \mathcal{Z})$ is shared across all agents. 
Together with discrete actions, this process forms local discrete sequences $( ..., x^{i}_{t, 1}, ..., x^{i}_{t, K}, a^{i}_{t}, ... )$ of each agent.
Then the Perceiver ({\bf \emph{right}}) performs aggregation of joint discrete sequences of all agents $(x^{1}_{t, 1}, ..., x^{1}_{t, K}, a^{1}_{t}, ..., x^{n}_{t, 1}, ..., x^{n}_{t, K}, a^{n}_{t})$ independently at each timestep $t$, and inserts the aggregated global representations $(e_{t}^{1}, e_{t}^{2}, ..., e_{t}^{n})$ into original local discrete sequences respectively. The resulting sequences $( ..., x^{i}_{t, 1}, ..., x^{i}_{t, K}, a^{i}_{t}, e^{i}_{t}... )$ contain rich information between transitions in local dynamics and are fed into the shared Transformer ({\bf \emph{middle}}), which learns observation token predictions in an autoregressive manner. Predictions of individual reward $r_{t}^{i}$ and discount $\gamma_{t}^{i}$ at timestep $t$ are computed based on all historical {sequences} $(x^{i}_{\leq t, 1}, ..., x^{i}_{\leq t, K}, a^{i}_{\leq t}, e^{i}_{\leq t})$.}
\label{fig:model_architecture}
\end{center}
\vspace{-2.3em} 
\end{figure*}

\section{Methodologies}

Our approach comprises three typical parts: (\romannumeral1)~collecting experience by executing the policy, (\romannumeral2)~learning the world model from the collected experience, and (\romannumeral3)~learning the policy via imagination inside the world model. Throughout the process, the historical experiences stored in the replay buffer are used for training the world model only, while policies are learned from unlimited imagined trajectories from the world model. In the following, we first describe three core components of our world model in \S\ref{sec:discretization} and \S\ref{sec:dynamics}, and give an overview of the proposed world model in Fig.~\ref{fig:model_architecture}. Then we describe the policy-learning process inside the world model in \S\ref{sec:behaviour_learning}. The comprehensive details of the model architecture and hyperparameters are provided in \S\ref{appendix:model_details}.

\subsection{Discretizing Observation}\label{sec:discretization}
We consider {that} a trajectory $\tau^{i}$ of agent $i$ consists of $T$ local observations and actions, as
\begin{align*}
    \tau^{i} = (o^{i}_{1}, a^{i}_{1}, \ldots, o^{i}_{t}, a^{i}_{t}, \ldots, o^{i}_{T}, a^{i}_{T}).
\end{align*}
To utilize the expressive Transformer architecture, we need to express the trajectory into a discrete token sequence for modeling. Accounting for continuous observations, a prevalent but naive approach for discretization involves discretizing the scalar into one of $m$ fixed-width bins in each dimension independently \citep{janner2021tt}.
{Such discretization would encode the observation with more tokens when faced with a higher dimension of the observation, resulting in longer sequence as the Transformer's input. Thus it causes higher computational complexity of sequence modeling via the Transformer because of the quadratic complexity with respect to the sequence length.}
{In contrast, a discrete codebook of learned compact representations can concisely encode a observation with less tokens.}
To this end, we employ the idea from neural discrete representation learning \citep{oord17vqvae}, and learn a Vector Quantised-Variational AutoEncoder (VQ-VAE) to play a role that resembles the tokenizer in Natural Language Processing \citep{devlin19bert, brown20llmfewshot}. The VQ-VAE is composed of an encoder $E$, a decoder $D$, and a codebook $\mathcal{Z}$. We define the discrete codebook $\mathcal{Z} = \{z_{j}\}^{N}_{j=1} \subset \mathbb{R}^{n_z}$, where $N$ is the size of the codebook and $n_z$ is the dimension of codes.
The encoder $E$ takes an observation $o^i \in \mathbb{R}^{n_{\rm obs}}$ as input and outputs a $K$ $n_z$-dimensional latents $\hat{z}^i \in \mathbb{R}^{K \times n_z}$ reshaped from the direct outputs of encoder.
Subsequently, the tokens $\{ x^{i}_{k} \}^{K}_{k=1} \in \{ 0, 1, ..., N - 1 \}^{K}$ for representing $o^i$ is obtained by a nearest neighbour look-up using the codebook $\mathcal{Z}$ where $x_{k}^{i} = \mathop{\arg\min}_{j} \| \hat{z}^{i}_{k} - z_j \|$.
Then the decoder $D: \{0, 1, ..., N - 1\}^{K} \rightarrow \mathbb{R}^{n_{\rm obs}}$ converts $K$ tokens back into an reconstructed observation $\hat{o}^i$.
By learning this discrete codebook, we compress the redundant information via a succinct sequence of tokens, which helps improve sequence modeling. See \S\ref{sec:ablation} for a discussion.

\subsection{Modeling Local Dynamics with Global Representations}\label{sec:dynamics}
Here, we consider discrete actions like those in SMAC, and the continuous actions can also be discretized by splitting the value in each dimension into fixed bins \citep{janner2021tt, 2022rt1}.
Therefore, a trajectory $\tau^{i}$ of agent $i$ can be treated as a sequence of tokens,
\begin{equation}
\label{equ:discrete_sequence}
    \tau^{i} = (\ldots, o_t^i, a_t^i, \ldots) = (\ldots, x_{t, 1}^{i}, x_{t, 2}^{i}, \ldots, x_{t, K}^{i}, a_{t}^{i}, \ldots)
\end{equation}
where $x_{t, j}^{i}$ is the $j$-th token of the observation of agent $i$ at timestep $t$.
Given arbitrary sequences of observation and action tokens in Eq.~\ref{equ:discrete_sequence}, we try to learn over discrete multimodal tokens.

The world model consists of a tokenizer to discrete the local observation, a Transformer to learn the local dynamics, an agent-wise representation aggregation module, and predictors for the reward and discount.
The Transformer $\phi$ predicts the future local observation $\{\hat{x}_{t+1, j}^{i}\}^{K}_{j=1}$, the future individual reward $\hat{r}_{t}^{i}$ and discount $\hat{\gamma}_{t}^{i}$, based on the agent's individual historical observation-action history $(x_{\leq t, \boldsymbol{\cdot}}^{i}, a_{\leq t}^{i})$ and aggregated global {features} $e_{t}^{i}$ of the agent. 
The modules are shown in Eqs.~\ref{equ:transition}--\ref{equ:aggregation}.
\begin{alignat}{2}
    & \text{Transition:} &\hat{x}_{t+1, \boldsymbol{\cdot}}^{i} &\sim p_{\phi} (\hat{x}_{t+1, \boldsymbol{\cdot}}^{i} | x_{\leq t, \boldsymbol{\cdot}}^{i}, a_{\leq t}^{i}, e_{\leq t}^{i}) \text{ with } \hat{x}_{t+1, k}^{i} \sim p_{\phi} (\hat{x}_{t+1, k}^{i} | x_{\leq t, \boldsymbol{\cdot}}^{i}, a_{\leq t}^{i}, e_{\leq t}^{i}, x_{t+1, < k}^{i}) \label{equ:transition} \\
    & \text{Reward:} &\hat{r}_{t}^{i} &\sim p_{\phi} (\hat{r}_{t}^{i} | x_{\leq t, \boldsymbol{\cdot}}^{i}, a_{\leq t}^{i}, e_{\leq t}^{i}) \label{equ:reward} \\
    & \text{Discount:} &\hat{\gamma}_{t}^{i} &\sim p_{\phi} (\hat{\gamma}_{t}^{i} | x_{\leq t, \boldsymbol{\cdot}}^{i}, a_{\leq t}^{i}, e_{\leq t}^{i}) \label{equ:discount} \\
    & \text{Aggregation:} &(e_{t}^{1}&, e_{t}^{2}, ..., e_{t}^{n}) = f_{\theta} (x_{t, 1}^{1}, x_{t, 2}^{1}, ..., x_{t, K}^{1}, a_{t}^{1}, ..., x_{t, 1}^{n}, x_{t, 2}^{n}, ..., x_{t, K}^{n}, a_{t}^{n}) \label{equ:aggregation}
\end{alignat}

\textbf{Transition Prediction.}
In the transition prediction in Eq.~\ref{equ:transition}, the $k$-th observation token is additionally conditioned on the tokens that were already predicted $x_{t+1, < k}^{i} \triangleq (x_{t+1, 1}^{i}, x_{t+1, 2}^{i}, ..., x_{t+1, k-1}^{i})$, ensuring the autoregressive token prediction to facilitate modeling over the trajectory sequence. 
Inter-step auto regression is as intuitive as predicting the future based on all information in the past while intra-step auto regression can be interpreted as learning how to compose the language provided by VQ-VAE to correctly express the observation within a certain timestep, since the tokens for encoding observations can be viewed as a special inner language like the human's.

\textbf{Discount and Reward Prediction.}
The discount predictor outputs a Bernoulli likelihood and lets us estimate the probability of an individual agent's episode ending when learning behaviors from model predictions. 
And we simply adopt a smooth L1 loss for training the prediction of reward.

\textbf{Agent-wise Aggregation.}
Due to the partial environment, the non-stationarity issue stems from the sophisticated agent-wise inter-dependency on local observations generation.
To address it, we introduce a Perceiver \citep{jaegle2021perceiver} to perform agent-wise representation aggregation which plays a similar role to communication.
To sustain the decentralized manner in transition prediction, we hope every agent can possess its own inner perception of the whole situation. Nonetheless, with discrete representation for local observation, the observation-action pair of agent $i$ at timestep $t$ is projected into a sequence $(x_{t, 1}^{i}, x_{t, 2}^{i}, ..., x_{t, K}^{i}, a_{t}^{i})$ of length $K+1$. It leads to a joint observation-action sequence of length $n(K + 1)$ at a timestep, which linearly scales with the number of agents.

A naive approach for extracting aggregated {features} for each agent is using self-attention \citep{Egorov22mamba, liu2024mazero} which takes as input this sequence of length $n(K+1)$ and outputs a sequence of the same length containing aggregated features of all agents, described as
\begin{align*}
    (x_{t, 1}^{1}, ..., x_{t, K}^{1}, a_{t}^{1}, ..., x_{t, 1}^{n}, ..., x_{t, K}^{n}, a_{t}^{n}) \xrightarrow[\text{Aggregating}]{\text{Self-Attention}} (e_{t, 1}^{1}, ..., e_{t, K}^{1}, e_{t, K + 1}^{1}, ..., e_{t, 1}^{n}, ..., e_{t, K}^{n}, e_{t,K+1}^{n}).
\end{align*}
where $e_{t,j}^{i}$ is the $j$-th aggregated feature for agent $i$ at timestep $t$. However, when composing the informative sequence of local trajectories by inserting these aggregated features into the sequence of length $H(K+1)$ in Eq.~\ref{equ:discrete_sequence}, the length of the local sequence involving aggregated features would be twice as long, i.e., $2H(K+1)$. Due to the quadratic computational complexity of Transformer, it may hinder the efficient sequence modeling over this sequence.

To this end, we choose the Perceiver as the agent-wise representation aggregation module, which excels at dealing with the case that the size of inputs scales linearly and then generates a compact output sequence. Equipped with a flexible querying mechanism and self-attention mechanism, the Perceiver aggregates the joint representation sequence $(x_{t, 1}^{1}, x_{t, 2}^{1}, ..., x_{t, K}^{1}, a_{t}^{1}, ..., x_{t, 1}^{n}, x_{t, 2}^{n}, ..., x_{t, K}^{n}, a_{t}^{n})$ of length $n(K + 1)$ into a sequence of $n$ features $(e_{t}^{1}, e_{t}^{2}, ..., e_{t}^{n})$,
\begin{align*}
    (x_{t, 1}^{1}, ..., x_{t, K}^{1}, a_{t}^{1}, ..., x_{t, 1}^{n}, ..., x_{t, K}^{n}, a_{t}^{n}) \xrightarrow[\text{Aggregating}]{\text{Perceiver}} (e_{t}^{1}, e_{t}^{2}, ..., e_{t}^{n})
\end{align*}
where each feature $e_{t}^{i}$ serves as an intrinsic global abstraction of the environmental contexts perceived from agent $i$'s viewpoint. By introducing Perceiver, we provide a feasible solution for reducing the modeling complexity when using transformer-based local dynamics.

\textbf{Overall Learning Objective.} The world model $\phi$ is trained with trajectory segments of a fixed horizon $H$ sampled from the replay buffer $\mathcal{D}$ in a self-supervised manner. The transition predictor, discount predictor, and reward predictor are optimized to maximize the log-likelihood of their corresponding targets:
\begin{align}\label{eq:loss}
    \mathcal{L}_{\rm Dyn} (\phi, \theta) =  \mathbb{E}_{i \sim \mathcal{N}} \mathbb{E}_{\tau^i \sim \mathcal{D}} \Bigl[ \sum\nolimits_{t=1}^{H} &- \underbrace{\log p_{\phi} (r_{t}^{i} | x_{\leq t, \boldsymbol{\cdot}}^{i}, a_{\leq t}^{i}, e_{t}^{i})}_{\text{reward loss}} - \underbrace{\log p_{\phi} (\gamma_{t}^{i} | x_{\leq t, \boldsymbol{\cdot}}^{i}, a_{\leq t}^{i}, e_{t}^{i})}_{\text{discount loss}} \nonumber \\
    & - \underbrace{\Bigl( \sum\nolimits_{k=1}^{K} \log p_{\phi} (x_{t+1, k}^{i} | x_{\leq t, \boldsymbol{\cdot}}^{i}, a_{\leq t}^{i}, e_{t}^{i}, x_{t+1, < k}^{i}) \Bigr)}_{\text{transition loss}} \Bigr] \\
    \mbox{where } (e_{t}^{1}, e_{t}^{2}, ..., e_{t}^{n}) = f_{\theta} &(x_{t, 1}^{1}, x_{t, 2}^{1}, ..., x_{t, K}^{1}, a_{t}^{1}, ..., x_{t, 1}^{n}, x_{t, 2}^{n}, ..., x_{t, K}^{n}, a_{t}^{n}), \forall t.
    \nonumber
\end{align}
We jointly minimize this loss function in Eq.~\ref{eq:loss} with respect to the model parameters of local dynamics (i.e., $\phi$) and global representation (i.e., $\theta$) using the Adam optimizer \citep{kingma14adam}.

\begin{figure*}[t]
\vskip -0.05in
\begin{center}
\centerline{
\includegraphics[width=1.0\columnwidth]{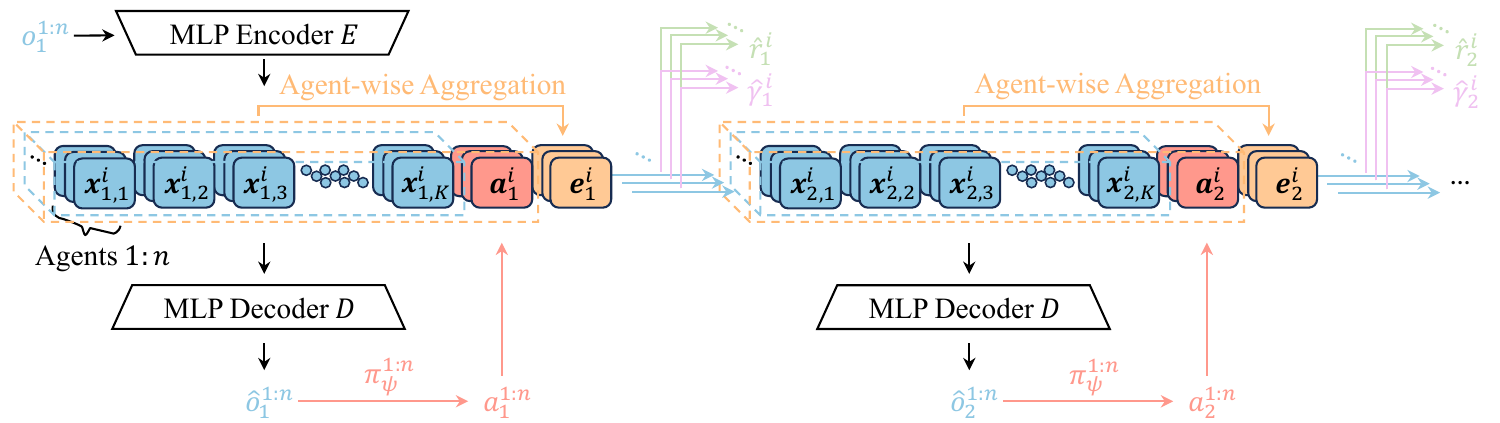}}
\vspace{-0.5em}
\caption{{\bf Imagination procedure in MARIE.}
We unroll the imagination of all agents $\{1, ..., n\}$ in parallel.
Initially, each agent's observation is derived from a joint observation sampled from a replay buffer.
A policy, depicted in \textcolor[rgb]{1.0, 0.596, 0.549}{red} arrows, generates actions based on reconstructed observations.
Then, the Perceiver integrates joint actions and observations into global representations from each agent, appending them to each agent's local sequence.
The Transformer then predicts individual rewards and discounts, depicted by \textcolor[rgb]{0.773, 0.878, 0.705}{green} and \textcolor[rgb]{0.941, 0.757, 0.945}{purple} arrows respectively, while generating next observation tokens for each agent in an autoregressive manner, shown by \textcolor[rgb]{0.553, 0.780, 0.890}{blue} arrows.
This parallel imagination iterates for $H$ steps. The policies $\pi^{1:n}_{\psi}$ are exclusively trained using imagined trajectories.
}
\label{fig:imagination}
\end{center}
\vspace{-2.3em}
\end{figure*}

\subsection{Learning Behaviours in Imagination}\label{sec:behaviour_learning}
We utilize the Actor-Critic framework to learn the behavior of each agent, where the actor and critic are parameterized by $\psi$ and $\xi$, respectively.
In the following, we take agent $i$ as an exemplar case for clarity and omit the superscript for denoting the index of the agent to avoid potential confusion.
{By benefiting} from the shared local dynamics, the local trajectories of all agents are imagined in parallel, as illustrated in Fig.~\ref{fig:imagination}.
At timestep $t$, the actor takes a reconstructed observation $\hat{o}_{t}$ as input, and samples an action $a_{t} \sim \pi_{\psi}(a_{t} | \hat{o}_t)$. The world model then predicts the individual reward $\hat{r}_{t}$, individual discount $\hat{\gamma}_t$ and next local observation $\hat{o}_{t+1}$. Starting from initial observations sampled from the replay buffer, this imagination procedure is rolled out for $H$ steps.
To stimulate long-horizon behavior learning, the critic accounts for rewards beyond the fixed imagination horizon and estimates the individual expected return $V_{\xi}(\hat{o}_{t}) \simeq \mathbb{E}_{\pi_{\psi}}[ \sum_{l \geq t} \gamma^{l - t} \hat{r}_{l} ]$. 

In our approach, we train the actor and critic in a MAPPO-like \citep{yu2022mappo} manner.
Unlike other CTDE model-free approaches that require a global oracle state from the environment, we cannot obtain the oracle state from the world model, and only the predicted observations of each agent are available. To approximate the oracle information in critic training, we enhance each agent's critic with the capability to access the observations of other agents.
Since the actor and critic only rely on the reconstructed observations, decoupling from the inner hidden states of the Transformer-based world model, we allow fast inference in the environment without the participation of the world model.
It is important for the deployment of policies learned with data-efficient imagination in real-world applications.
$\lambda$-target in Dreamer \citep{hafner2019dreamer} is used to updated the value function.
The details of behavior learning objectives and algorithmic description of MARIE are presented in \S\ref{appendix:behaviour_learning} and \S\ref{appendix:algo}, respectively.

\section{Experiments}\label{sec:experiments}
We consider the most common benchmark -- StarCraftII Multi-Agent Challenge (SMAC) \citep{samvelyan19smac} for evaluating our method.
To highlight the sample efficiency brought by model-based imagination, we adopt a low data regime that resembles a similar setting in single-agent Atari domain \citep{Kaiser2020SimPLe}. Additional experiment results on MAMujoco \citep{peng2021facmac} (i.e., continuous action space case) is provided in \S\ref{appendix:mamujoco_exp}.

\subsection{Experiment Setup and Evaluations}\label{sec:main_exp}
{\bf StarCraftII Multi-Agent Challenge.} SMAC \citep{samvelyan19smac}, a suite of cooperative multi-agent environments based on StarCraft II, consists of a set of StarCraft II scenarios. Each scenario depicts a confrontation between two armies of units, one of which is controlled by the built-in game AI and the other by our algorithm.
The initial position, number, and type of units in each army varies from scenario to scenario, as does the presence or absence of elevated or impassable terrain. And the goal is to win the game within the pre-specified time limit. SMAC emphasizes mastering micromanagement techniques across multiple agents to achieve effective coordination and overcome adversaries. This necessitates both sufficient exploration and appropriate credit assignment for each agent's action. Another notable property of SMAC is that not all actions are accessible during decision-making of each agent, which requires world models to possess an in-depth comprehension of the underlying game mechanics so as to consistently provide valid available action mask estimation within the imagination horizon. Thus, in this benchmark, we additionally add one more head for the prediction of available action mask. During the imagination of MARIE, the available action mask is estimated by this head, instead of being generated manually according to the meaning of each element in the reconstructed observation. The latter introduces too much prior knowledge about StarCraft and can be considered as benchmark hacking.

{\bf Experimental Setup.} 
We choose 13 representative scenarios from SMAC that includes three levels of difficulty -- \emph{Easy}, \emph{Hard}, and \emph{SuperHard}.
Specific chosen scenarios can be found in Table~\ref{table:main_exp}. In terms of different levels of difficulty, we adopt a similar setting akin to that in \citep{Egorov22mamba} and restrict the number of samples from the real environment to 100k for \emph{Easy} scenarios, 200k for \emph{Hard} scenarios and 400k for \emph{SuperHard} scenarios, to establish a low data regime in SMAC.
We compare MARIE with three strong model-free baselines -- MAPPO \citep{yu2022mappo}, QMIX \citep{rashid18qmix} and QPLEX \citep{wang2021qplex}, and two strong model-based baselines with the same policy learning paradigm as ours -- MBVD \citep{xu2022mbvd} and MAMBA \citep{Egorov22mamba} on SMAC benchmark. Specially, as a multi-agent variant of DreamerV2 \citep{hafner2020dreamerv2}, MAMBA achieves powerful sample efficiency in various SMAC scenarios via learning in imagination. For each random seed, we compute the win rate across 10 evaluation games at fixed intervals of environmental steps.
The hyperparameters of MARIE and other baselines are listed in \S\ref{app:baseline_hyper} and \S\ref{appendix:marie_hyper}. Particularly, the hyperparameters of model-free baselines in low data regime are directly referred to \citet{Egorov22mamba} and \citet{liu2024mazero}.

\begin{figure*}[h!]
\vskip -0.05in
\begin{center}
\centerline{
\includegraphics[width=1.0\columnwidth]{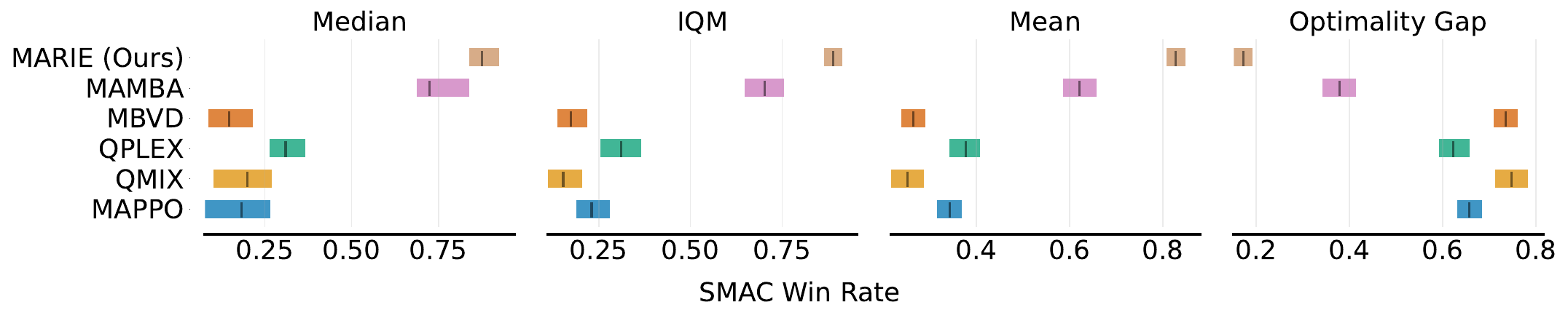}
}
\vspace{-0.5em}
\caption{{\bf MARIE exhibits statistically significant improvements on SMAC compared to the baselines}. We report mean, median, and inter-quartile mean win rate, computed with stratified bootstrap confidence intervals. By default, all algorithms are evaluated using 4 runs.}
\label{fig:iqm}
\end{center}
\vspace{-1.8em}
\end{figure*}

\begin{figure*}[t!]
\vskip -0.05in
\begin{center}
\centerline{
\includegraphics[width=1.0\columnwidth]{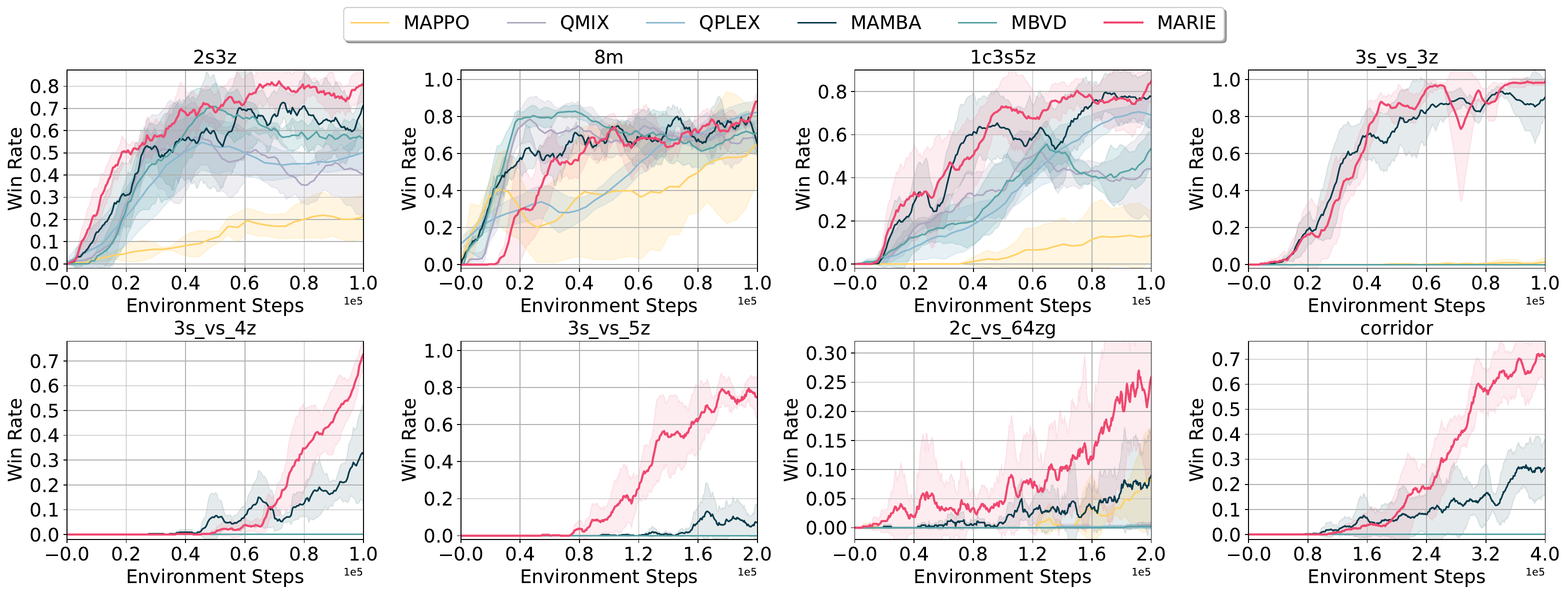}}
\vspace{-0.5em}
\caption{
{\bf The gap in sample efficiency and final performance between MARIE and the chosen baselines becomes more pronounced with increasing difficulty of scenarios.}
Curves of evaluation win rate for methods in 8 chosen SMAC maps. See Table~\ref{table:main_exp} for win rates. Y axis: win rate; X axis:  number of steps taken in the real environment.}
\label{fig:main_exp_result}
\end{center}
\vspace{-3em}
\end{figure*}

\begin{table}[tb!]
\centering
\caption{{\bf MARIE outperforms every considered baselines on all task with more than 3 agents.} We report mean evaluation win rate and standard deviation on 13 SMAC maps for different methods over 4 random seeds. And we bold the values of the maximum and highlight them with blue color.}
\vspace{1em}
\label{table:main_exp}
\resizebox{\columnwidth}{!}{%
\begin{tabular}{ccccccccc}
\toprule
\multirow{3}{*}{Maps} & \multirow{3}{*}{Difficulty}     & \multirow{3}{*}{Steps} & \multicolumn{6}{c}{Methods}                                                                      \\ \cmidrule(lr){4-9} 
                      &                                 &                        & MARIE             & MAMBA               & MAPPO      & QMIX       & QPLEX          & MBVD                   \\ 
                      &                                 &                        & (Ours)             & \citep{Egorov22mamba}               & \citep{yu2022mappo}      & \citep{rashid18qmix}       & \citep{wang2021qplex}          & \citep{xu2022mbvd}                   \\ \midrule
1c3s5z                & \multirow{10}{*}{\textit{Easy}} & \multirow{10}{*}{100K} & \colorbox{mine}{\textbf{85.0}\scr{(9.4)}}  & 77.7\scr{(15.3)}         & 18.4\scr{(11.0)} & 43.6\scr{(29.2)} & 68.3\scr{(7.4)}         & 60.9\scr{(11.4)}             \\
2m\_vs\_1z            &                                 &                        & \colorbox{mine}{\textbf{95.5}\scr{(7.9)}} & \colorbox{mine}{\textbf{95.5}\scr{(2.3)}}           & 86.7\scr{(3.2)}  & 70.3\scr{(14.8)} & 84.8\scr{(10.8)}          & 36.7\scr{(24.5)}             \\
2s\_vs\_1sc           &                                 &                        & 96.9\scr{(7.1)} & 95.0\scr{(7.1)}           & \colorbox{mine}{\textbf{100.0}\scr{(0.0)}} & 0.0\scr{(0.0)}   & 15.7\scr{(19.5)}          & 8.7\scr{(14.8)}              \\
2s3z                  &                                 &                        & \colorbox{mine}{\textbf{80.5}\scr{(9.3)}}  & 71.6\scr{(12.7)}          & 31.2\scr{(12.9)} & 37.7\scr{(15.5)} & 50.2\scr{(8.4)}           & 53.4\scr{(4.1)}              \\
3m                    &                                 &                        & \colorbox{mine}{\textbf{99.5}\scr{(0.4)}}  & 87.7\scr{(7.1)}           & 80.5\scr{(12.8)} & 54.4\scr{(22.7)} & 88.7\scr{(6.9)}           & 73.9\scr{(6.9)}              \\
3s\_vs\_3z            &                                 &                        & \colorbox{mine}{\textbf{98.9}\scr{(1.5)}}  & 89.3\scr{(10.1)}          & 1.2\scr{(1.3)}   & 0.0\scr{(0.0)}   & 0.0\scr{(0.0)}            & 0.0\scr{(0.0)}               \\
3s\_vs\_4z            &                                 &                        & \colorbox{mine}{\textbf{73.0}\scr{(6.2)}}  & 29.3\scr{(12.3)}          & 0.0\scr{(0.0)}   & 0.0\scr{(0.0)}   & 0.0\scr{(0.0)}            & 0.0\scr{(0.0)}               \\
8m                    &                                 &                        & \colorbox{mine}{\textbf{88.0}\scr{(3.9)}}  & 65.0\scr{(7.7)}           & 70.3\scr{(19.5)} & 69.5\scr{(12.8)} & 83.4\scr{(6.4)}           & 74.7\scr{(9.7)}              \\
MMM                   &                                 &                        & \colorbox{mine}{\textbf{87.6}\scr{(3.0)}}  & 50.2\scr{(27.6)}          & 5.5\scr{(4.5)}   & 31.1\scr{(17.3)} & 69.3\scr{(35.1)}          & 20.5\scr{(2.1)}              \\
so\_many\_baneling    &                                 &                        & \colorbox{mine}{\textbf{94.8}\scr{(5.9)}}  & 91.6\scr{(4.1)}           & 43.8\scr{(15.0)} & 20.0\scr{(8.9)}  & 32.2\scr{(6.1)}           & 15.0\scr{(10.4)}             \\ \midrule
3s\_vs\_5z            & \multirow{2}{*}{\textit{Hard}}  & \multirow{2}{*}{200K}  & \colorbox{mine}{\textbf{78.4}\scr{(11.2)}} & 13.4\scr{(14.0)}          & 0.0\scr{(0.0)}   & 0.0\scr{(0.0)}   & 0.0\scr{(0.0)}            & 0.0\scr{(0.0)}               \\
2c\_vs\_64zg          &                                 &                        & \colorbox{mine}{\textbf{25.9}\scr{(14.3)}} & 9.8\scr{(8.7)}            & 7.8\scr{(10.2)}  & 0.5\scr{(0.5)}   & 0.1\scr{(0.1)}            & 0.2\scr{(0.4)}               \\ \midrule
corridor              & \textit{SuperHard}              & 400K                   & \colorbox{mine}{\textbf{71.0}\scr{(13.8)}} & 26.5\scr{(15.2)}          & 0.4\scr{(0.7)}   & 0.0\scr{(0.0)}   & 0.0\scr{(0.0)}            & 0.0\scr{(0.0)}               \\ \midrule
\multicolumn{3}{c}{Model Param Size} & 55.35 MB & 23.72 MB & 0.24 MB & 0.19 MB & 0.23 MB & 0.84 MB \\ \bottomrule
\end{tabular}%
}
\vspace{-1em}
\end{table}

\subsection{Main Results on SMAC}\label{sec:smac_results}
{{\bf MARIE achieves superior sample efficiency and final win rate to model-based and model-free multi-agent baselines.}}
{\citet{agarwal2021rliable} discuss the limitations of mean and median scores, and show that substantial discrepancies arise between standard point estimates and interval estimates in RL benchmarks.}
{To deliver a rigorous statistical evaluation, we summarize in Figure~\ref{fig:iqm} the win rate with stratified bootstrap confidence intervals for mean, median, and inter-quartile mean (IQM)}.
Overall, we find MARIE achieves significantly better sample efficiency and a higher win rate compared with other strong baselines. {Furthermore}, we report the detailed averaged win rates over four seeds in every chosen scenario in Table~\ref{table:main_exp} and provide additional learning curves of several chosen scenarios in Fig.~\ref{fig:main_exp_result}. 
As presented in Table~\ref{table:main_exp} and Fig.~\ref{fig:main_exp_result}, MARIE demonstrates superior performance and sample efficiency across almost all scenarios.
The improvements in sample efficiency and performance become particularly pronounced with increasing difficulty of scenarios, especially compared to MAMBA that adopts RSSM as the backbone for the world model.
We attribute such results to the model capability of the Transformer in local dynamics modeling and global feature aggregation.
Benefiting from more powerful strength in modeling sequences, the Transformer-based world model can generate more accurate and consistent imaginations than those relying on the recurrent backbone, which facilitates better policy learning within the imagination of the world model.
While the scenarios become harder, e.g. \emph{3s\_vs\_5z}, our world model can address the challenge of learning more intricate underlying dynamics and further large quantities of accurate imaginations, thereby significantly outperforming other baselines on these scenarios.
Moreover, a special scenario \emph{2c\_vs\_64zg} deserves attention, which features only 2 agents but with a considerably large action space of up to 70 discrete actions for each agent.
Although the performance of MARIE in \emph{2c\_vs\_64zg} suffers a relative large variance due to the overly large action space, MARIE achieves a remarkably non-trivial mean win rate just via learning in the imagination.
Note that it is easy for the world model to generate ridiculous estimated available action masks without understanding the mechanics behind this scenario, further leading to invalid or even erroneous policy learning in the imaginations of the world model. The performance gap on \emph{2c\_vs\_64zg} proves that our Transformer-based world model has higher prediction accuracy and a deeper understanding of the underlying mechanics.

{On the other hand, MARIE contains more than twice the number of parameters as MAMBA, which may partially account for its more pronounced improvement. This observation further implies that the learning of world models may benefit from scaling laws, albeit at the cost of increasing computational overhead and longer training time.}

\subsection{Ablation Studies}\label{sec:ablation}
\begin{figure*}[t!]
\vskip -0.05in
\begin{center}
\centerline{
\includegraphics[width=1.0\columnwidth]{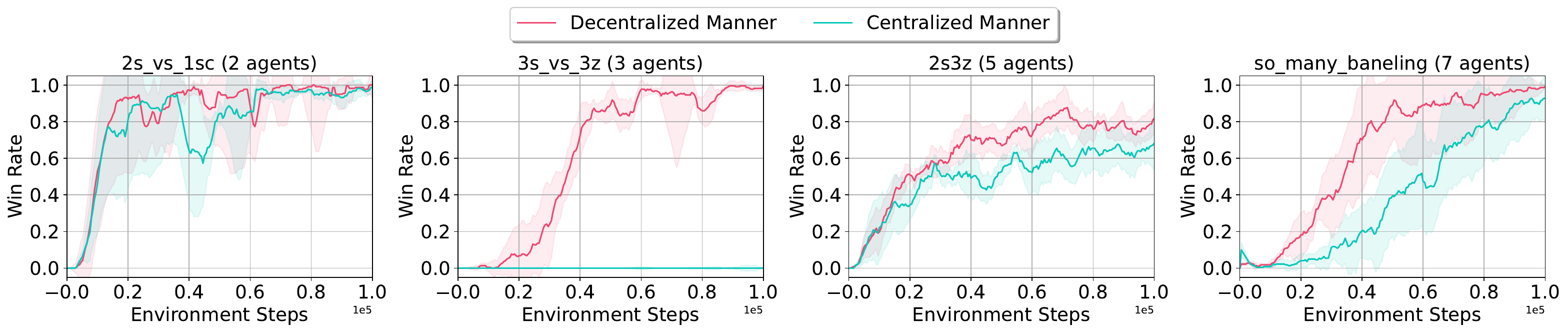}}
\vspace{-0.5em}
\caption{{\bf Decentralized dynamics modeling allows for better scaling to environments with growing number of agents.} We conduct an ablation study on what manner to integrate into the design of the world model. \emph{Decentralized} Manner denotes the standard implementation of MARIE, while \emph{Centralized} Manner denotes that the world model is designed for learning the joint dynamics of all agents over the joint trajectory.
{Sample efficiency of the \emph{centralized} variant encounters a significant drop due to the scalability issue while MARIE is robust to scenarios with various number of agents.}
}
\vspace{-2em}
\label{fig:central_ablation_result}
\end{center}
\end{figure*}

{\bf Incorporating CTDE principle with the design of the world model makes MARIE scalable and robust to different number of agents.}
We compare our method with a \emph{centralized} variant of our method, wherein the world model learns the joint dynamics of all agents together over the joint trajectory $\tau = (\ldots, o_{t}^{1}, o_{t}^{2}, \ldots, o_{t}^{n}, a_{t}^{1}, a_{t}^{2}, \ldots, a_{t}^{n}, \ldots)$.
Given that $\tau$ already contains the joint observations and actions, we disable the aggregation module in this \emph{centralized} variant. As illustrated in Figure \ref{fig:central_ablation_result}, our comparisons span scenarios involving 2 to 7 agents.
When the number of agents is small enough, reducing the multi-agent system to a single-agent one over the joint observation and action space would not cause a prominent scalability issue, as indicated by the result in \emph{2s\_vs\_1sc}.
However, the scalability issue is exacerbated by a growing number of agents.
In scenarios featuring more than 3 agents, the sample efficiency of the \emph{centralized} variant encounters a significant drop, suffering from the exponential surge in spatial complexity of the joint observation-action space.
Furthermore, with equal prediction horizons, the parameter amounts in the \emph{centralized} variant is increased by a factor of 4 or larger. And to achieve the same number of environment steps, the \emph{centralized} variant demands over twice the original computational time.
Instead, with decentralized local dynamics and aggregated global features, MARIE delivers stable and superior sample efficiency.

\begin{figure*}[t!]
\vskip -0.05in
\begin{center}
\centerline{
\includegraphics[width=1.0\columnwidth]{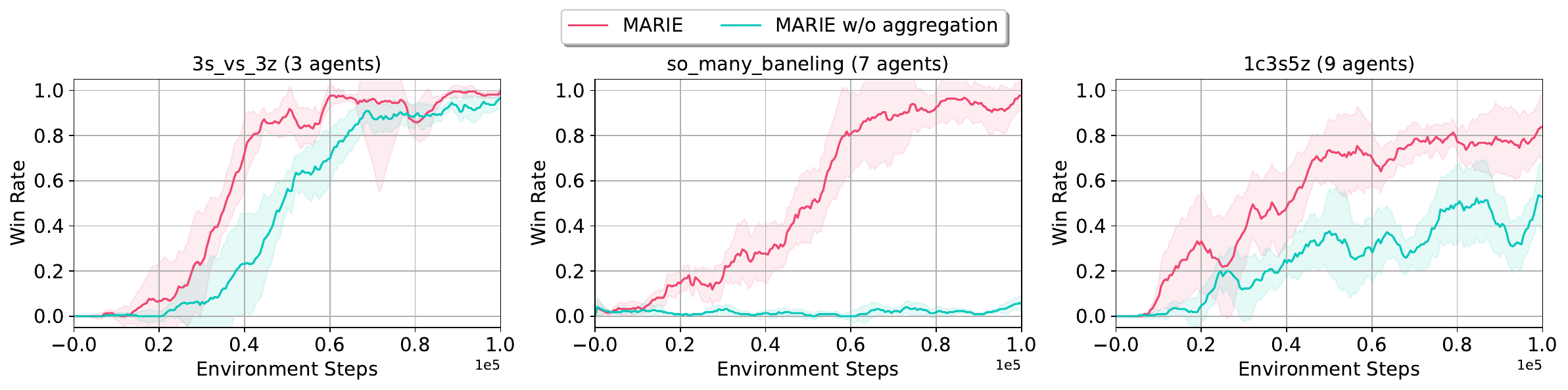}}
\vspace{-0.5em}
\caption{
{\bf Local dynamics struggles to infer accurate future local observations without agent-wise aggregation.}
We plot the comparison results between MARIE with and without the usage of the aggregation module.}
\label{fig:perceiver_ablation_result}
\end{center}
\vspace{-2em}
\end{figure*}

{\bf Agent-wise aggregation helps MARIE capture the sophisticated inter-dependency on the generation of each agent's local observation.} 
To study the influence of agent-wise aggregation, we conduct ablation experiments on the aggregation module over scenarios where the number of agents gradually increases. 
As shown in Fig.~\ref{fig:perceiver_ablation_result}, in the 3-agents scenario (e.g., \emph{3s\_vs\_3z}), the correlation among each agent's local observation tends to be negligible. Therefore, the nearly independent generation of each agent's local observation without any aggregated global feature still leads to performance comparable to that of standard implementation.
But as more agents get involved, the inter-dependency becomes dominant. Lacking the global features derived from agent-wise aggregation, the shared Transformer struggles to infer accurate future local observations, thus hindering policy learning in the imaginations of the world model and resulting in notable degradation in the win rate evaluation.

\begin{figure*}[t!]
\vskip -0.05in
\begin{center}
\centerline{
\includegraphics[width=1.0\columnwidth]{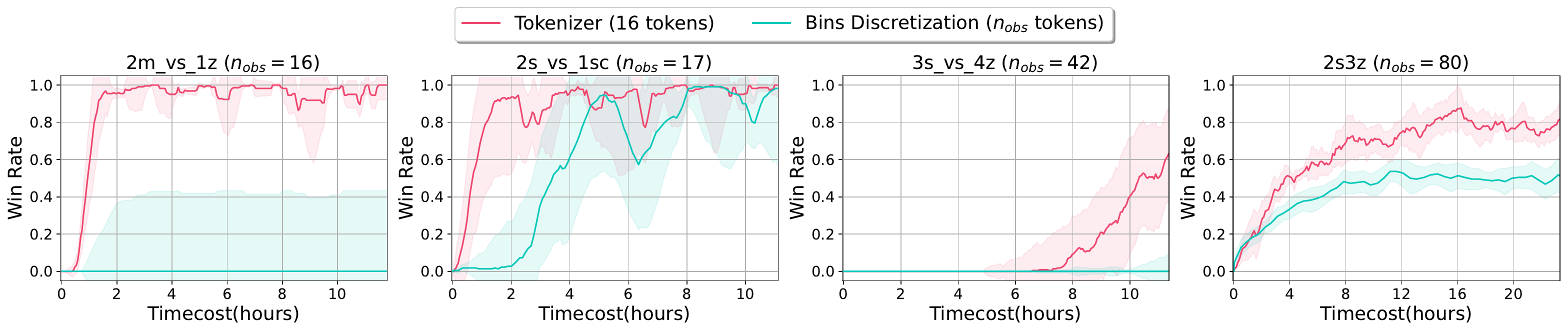}}
\vspace{-0.5em}
\caption{{\bf Vector quantization is more effective than Bins Discretization for encoding observations with discrete tokens.}
We conduct an ablation study on the type of discretization for local observations. \emph{Tokenizer} denotes the standard implementation of MARIE;  \emph{Bins Discretization} denotes the variant of MARIE where the $n_{obs}$-dimensional observation discretization is performed by projecting the value into one of $m$ fixed-width bins in each dimension independently. X-axis: cumulative run time of algorithms in the same platform.
{VQ-VAE encapsulates local observations within a succinct sequence of tokens, computationally efficiently promoting the learning of the Transformer-based world model.}
}
\label{fig:bins_ablation_result}
\end{center}
\vspace{-2em}
\end{figure*}

{\bf VQ-VAE encapsulates local observations within a succinct sequence of tokens, promoting the learning of the Transformer-based world model and effectively improving algorithm performance.}
Compared to VQ-VAE that discretizes each observation to $K$ tokens from $ \mathcal{Z}$, perhaps a more naive tokenizer is projecting the value in each dimension into one of $m$ fixed-width bins \citep{janner2021tt}, resulting in a $n_{obs}$-long token sequence for each observation, which we term \emph{Bins Discretization}. We set the number of bins $m$ equal to the size of codebook $|\mathcal{Z}|$ and compare these two types of tokenizers in different environments with various $n_{\rm obs}$.
As shown in Fig.~\ref{fig:bins_ablation_result}, the performance of the two tokenizers are comparable only in \emph{2s\_vs\_1sc} where $n_{\rm obs}$ is close to 16. Even worse, \emph{Bins Discretization} experiences a pronounced decline as $n_{\rm obs}$ increases in more complex environments (e.g., \emph{3s\_vs\_4z}) under identical training durations.
We hypothesize that for a single local observation, a $n_{\rm obs}$-token-long verbose sequence yielded by \emph{Bins Discretization} contains more redundant information compared to VQ-VAE that learns a more compact tokenizer through reconstruction.
This not only renders the token sequences of \emph{Bins Discretization} obscure and challenging to comprehend, but also results in an increase in model parameter amounts, being more computationally costly. Due to these two factors, \emph{Bins Discretization} exhibits a notably slow convergence.
Meanwhile, the result in \emph{2m\_vs\_1z} indicates \emph{Bins Discretization} may ignore the correlation of different dimensions, which would be helpful in sequence modeling.

\subsection{Model Analysis}
\begin{figure*}[t!]
\begin{center}
\centerline{
\includegraphics[width=1.0\columnwidth]{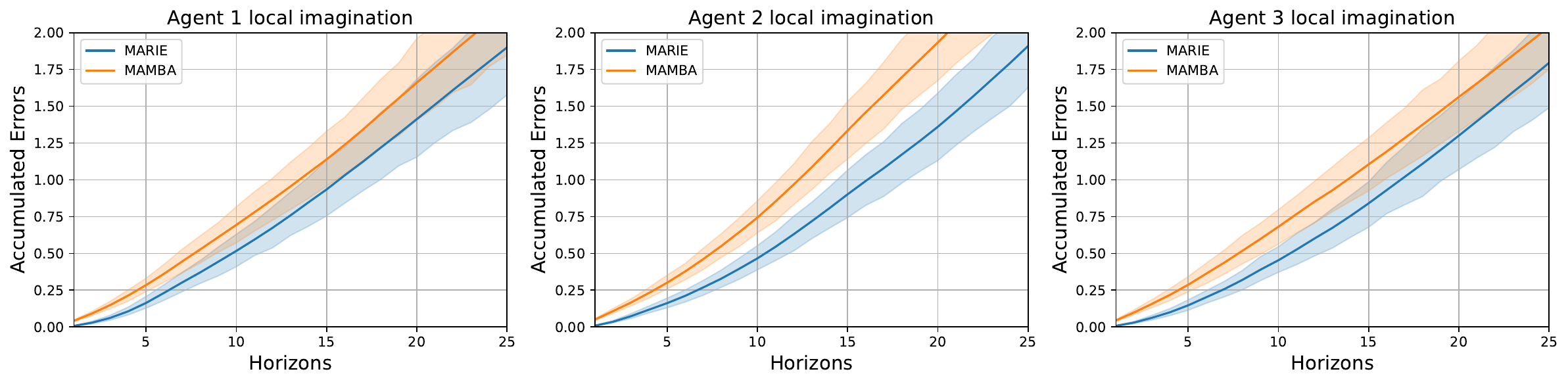}}
\vspace{-0.5em}
\caption{{\bf MARIE exhibits significantly lower compounding prediction error than MAMBA across all agents with respect to prediction horizon.} We compare the imagination accuracy of MARIE to that of MAMBA over the course of a planning horizon in \emph{3s\_vs\_5z} scenario.}
\label{fig:accumulated_errors}
\end{center}
\vspace{-2em}
\end{figure*}

\textbf{Error Accumulation.}
A quantitative evaluation of the model's accumulated error versus prediction horizon is provided in Fig.~\ref{fig:accumulated_errors}. Since learning the world model is tied to a progressively improving policy both in MARIE and MAMBA, we separately use their final policies to sample 10 episodes for fairness. We then compute $L_1$ errors per observation dimension between 1000 trajectory segments randomly sampled from these 20 episodes and their imagined counterpart. The result in Fig.~\ref{fig:accumulated_errors} suggests architecture differences play a large role in the world model’s long-horizon accuracy. This also provides additional evidence that policy learning can benefit from accurate long-term imaginations, explaining MARIE's notable performance in the \emph{3s\_vs\_5z} scenario.
More precisely, lower generalization error between the estimated dynamics and true dynamics brings a tighter bound between optimal policies derived from these two dynamics according to theoretical results \citep{janner2019mbpo}.

\textbf{Attention Patterns.}
During model prediction, we delve into the attention maps inside the shared Transformer and the cross attention maps in the Perceiver. Interestingly, we observe two distinct attention patterns involved in the local dynamics prediction. One exhibits a Markovian pattern wherein the observation prediction lays its focus mostly on the previous transition, while the other is regularly striated wherein the model attends to specific tokens in multiple prior transitions. During the agent-wise aggregation, we also identify two distinct patterns -- \emph{individuality} and \emph{commonality} among agents. Such diverse patterns in the Transformer and Perceiver may be pivotal for achieving accurate and consistent imaginations of the sophisticated local dynamics. We refer to \S\ref{appendix:attn_vis} for further details and visualization results.

\section{Conclusion and Limitation}
We have introduced a model-based multi-agent algorithm -- MARIE, which utilizes a shared Transformer as a local dynamic model and a Perceiver as a global agent-wise aggregation module to construct a world model within the multi-agent context. 
By providing long-term imaginations with policy learning, it significantly boosts the sample efficiency and improves final performance compared to state-of-the-art model-free methods and existing model-based methods with the same learning paradigm, in the low data regime.
{But it should be also noted that there are potential limitations on the current evaluation on the main experiment with 4 limited seeds, e.g., the limitations of mean and median scores \citep{agarwal2021rliable}. Thus, we also provide a standardized performance evaluation following the protocol provided by \citet{agarwal2021rliable} {in \S\ref{sec:smac_results} and \S\ref{appendix:rliable}}. To further deliver a rigorous statistical validation, evaluation with more seeds is definitely necessary.}
As the first Transformer-based multi-agent world model for sample-efficient policy learning, we open a new avenue for combining the powerful strength of the Transformer with sample-efficient MARL.
Considering the notorious sample inefficiency in multi-agent scenarios, it holds important promise for application in many realistic multi-robot systems, wherein collecting tremendous samples for optimal policy learning is costly and impractical due to safety.
While it has the great potential to {illuminate} the future towards achieving smarter multi-agent systems, there still exist limitations in MARIE. For instance, it would suffer from much slower inference speed when used with very long prediction horizons, due to the auto-regressive property. 


\bibliography{main}

\begin{thebibliography}{54}
\providecommand{\natexlab}[1]{#1}
\providecommand{\url}[1]{\texttt{#1}}
\expandafter\ifx\csname urlstyle\endcsname\relax
  \providecommand{\doi}[1]{doi: #1}\else
  \providecommand{\doi}{doi: \begingroup \urlstyle{rm}\Url}\fi

\bibitem[Agarwal et~al.(2021)Agarwal, Schwarzer, Castro, Courville, and Bellemare]{agarwal2021rliable}
Rishabh Agarwal, Max Schwarzer, Pablo~Samuel Castro, Aaron~C Courville, and Marc Bellemare.
\newblock Deep reinforcement learning at the edge of the statistical precipice.
\newblock \emph{Advances in neural information processing systems}, 34:\penalty0 29304--29320, 2021.

\bibitem[Brody et~al.(2022)Brody, Alon, and Yahav]{brody2021attentive}
Shaked Brody, Uri Alon, and Eran Yahav.
\newblock How attentive are graph attention networks?
\newblock In \emph{International Conference on Learning Representations}, 2022.
\newblock URL \url{https://openreview.net/forum?id=F72ximsx7C1}.

\bibitem[Brohan et~al.(2023)Brohan, Brown, Carbajal, Chebotar, Dabis, Finn, Gopalakrishnan, Hausman, Herzog, Hsu, Ibarz, Ichter, Irpan, Jackson, Jesmonth, Joshi, Julian, Kalashnikov, Kuang, Leal, Lee, Levine, Lu, Malla, Manjunath, Mordatch, Nachum, Parada, Peralta, Perez, Pertsch, Quiambao, Rao, Ryoo, Salazar, Sanketi, Sayed, Singh, Sontakke, Stone, Tan, Tran, Vanhoucke, Vega, Vuong, Xia, Xiao, Xu, Xu, Yu, and Zitkovich]{2022rt1}
Anthony Brohan, Noah Brown, Justice Carbajal, Yevgen Chebotar, Joseph Dabis, Chelsea Finn, Keerthana Gopalakrishnan, Karol Hausman, Alex Herzog, Jasmine Hsu, Julian Ibarz, Brian Ichter, Alex Irpan, Tomas Jackson, Sally Jesmonth, Nikhil Joshi, Ryan Julian, Dmitry Kalashnikov, Yuheng Kuang, Isabel Leal, Kuang-Huei Lee, Sergey Levine, Yao Lu, Utsav Malla, Deeksha Manjunath, Igor Mordatch, Ofir Nachum, Carolina Parada, Jodilyn Peralta, Emily Perez, Karl Pertsch, Jornell Quiambao, Kanishka Rao, Michael Ryoo, Grecia Salazar, Pannag Sanketi, Kevin Sayed, Jaspiar Singh, Sumedh Sontakke, Austin Stone, Clayton Tan, Huong Tran, Vincent Vanhoucke, Steve Vega, Quan Vuong, Fei Xia, Ted Xiao, Peng Xu, Sichun Xu, Tianhe Yu, and Brianna Zitkovich.
\newblock Rt-1: Robotics transformer for real-world control at scale.
\newblock In \emph{Robotics: Science and Systems (RSS)}, 2023.

\bibitem[Brown et~al.(2020)Brown, Mann, Ryder, Subbiah, Kaplan, Dhariwal, Neelakantan, Shyam, Sastry, Askell, Agarwal, Herbert-Voss, Krueger, Henighan, Child, Ramesh, Ziegler, Wu, Winter, Hesse, Chen, Sigler, Litwin, Gray, Chess, Clark, Berner, McCandlish, Radford, Sutskever, and Amodei]{brown20llmfewshot}
Tom Brown, Benjamin Mann, Nick Ryder, Melanie Subbiah, Jared~D Kaplan, Prafulla Dhariwal, Arvind Neelakantan, Pranav Shyam, Girish Sastry, Amanda Askell, Sandhini Agarwal, Ariel Herbert-Voss, Gretchen Krueger, Tom Henighan, Rewon Child, Aditya Ramesh, Daniel Ziegler, Jeffrey Wu, Clemens Winter, Chris Hesse, Mark Chen, Eric Sigler, Mateusz Litwin, Scott Gray, Benjamin Chess, Jack Clark, Christopher Berner, Sam McCandlish, Alec Radford, Ilya Sutskever, and Dario Amodei.
\newblock Language models are few-shot learners.
\newblock In \emph{Advances in Neural Information Processing Systems}, 2020.
\newblock URL \url{https://proceedings.neurips.cc/paper_files/paper/2020/file/1457c0d6bfcb4967418bfb8ac142f64a-Paper.pdf}.

\bibitem[Chen et~al.(2022)Chen, Wu, Yoon, and Ahn]{chen2022transdreamer}
Chang Chen, Yi-Fu Wu, Jaesik Yoon, and Sungjin Ahn.
\newblock Transdreamer: Reinforcement learning with transformer world models.
\newblock \emph{arXiv preprint arXiv:2202.09481}, 2022.

\bibitem[Devlin et~al.(2019)Devlin, Chang, Lee, and Toutanova]{devlin19bert}
Jacob Devlin, Ming{-}Wei Chang, Kenton Lee, and Kristina Toutanova.
\newblock {BERT:} pre-training of deep bidirectional transformers for language understanding.
\newblock In \emph{Proceedings of the 2019 Conference of the North American Chapter of the Association for Computational Linguistics: Human Language Technologies, {NAACL-HLT}}, 2019.
\newblock \doi{10.18653/V1/N19-1423}.
\newblock URL \url{https://doi.org/10.18653/v1/n19-1423}.

\bibitem[Egorov \& Shpilman(2022)Egorov and Shpilman]{Egorov22mamba}
Vladimir Egorov and Alexei Shpilman.
\newblock Scalable multi-agent model-based reinforcement learning.
\newblock In \emph{Proceedings of the 21st International Conference on Autonomous Agents and Multiagent Systems}, 2022.

\bibitem[Ellis et~al.(2023)Ellis, Cook, Moalla, Samvelyan, Sun, Mahajan, Foerster, and Whiteson]{ellis2023smacv2}
Benjamin Ellis, Jonathan Cook, Skander Moalla, Mikayel Samvelyan, Mingfei Sun, Anuj Mahajan, Jakob~Nicolaus Foerster, and Shimon Whiteson.
\newblock {SMAC}v2: An improved benchmark for cooperative multi-agent reinforcement learning.
\newblock In \emph{Thirty-seventh Conference on Neural Information Processing Systems Datasets and Benchmarks Track}, 2023.
\newblock URL \url{https://openreview.net/forum?id=5OjLGiJW3u}.

\bibitem[Foerster et~al.(2018)Foerster, Farquhar, Afouras, Nardelli, and Whiteson]{Foerster18coma}
Jakob~N. Foerster, Gregory Farquhar, Triantafyllos Afouras, Nantas Nardelli, and Shimon Whiteson.
\newblock Counterfactual multi-agent policy gradients.
\newblock In \emph{Proceedings of the Thirty-Second AAAI Conference on Artificial Intelligence}, 2018.

\bibitem[Ha \& Schmidhuber(2018)Ha and Schmidhuber]{david2018worldmodels}
David Ha and J\"{u}rgen Schmidhuber.
\newblock Recurrent world models facilitate policy evolution.
\newblock In \emph{Advances in Neural Information Processing Systems}, 2018.

\bibitem[Hafner et~al.(2019)Hafner, Lillicrap, Fischer, Villegas, Ha, Lee, and Davidson]{hafner19PlaNet}
Danijar Hafner, Timothy Lillicrap, Ian Fischer, Ruben Villegas, David Ha, Honglak Lee, and James Davidson.
\newblock Learning latent dynamics for planning from pixels.
\newblock In \emph{Proceedings of the 36th International Conference on Machine Learning}, Proceedings of Machine Learning Research. PMLR, 2019.

\bibitem[Hafner et~al.(2020)Hafner, Lillicrap, Ba, and Norouzi]{hafner2019dreamer}
Danijar Hafner, Timothy Lillicrap, Jimmy Ba, and Mohammad Norouzi.
\newblock Dream to control: Learning behaviors by latent imagination.
\newblock In \emph{International Conference on Learning Representations}, 2020.
\newblock URL \url{https://openreview.net/forum?id=S1lOTC4tDS}.

\bibitem[Hafner et~al.(2021)Hafner, Lillicrap, Norouzi, and Ba]{hafner2020dreamerv2}
Danijar Hafner, Timothy~P Lillicrap, Mohammad Norouzi, and Jimmy Ba.
\newblock Mastering atari with discrete world models.
\newblock In \emph{International Conference on Learning Representations}, 2021.
\newblock URL \url{https://openreview.net/forum?id=0oabwyZbOu}.

\bibitem[Hafner et~al.(2023)Hafner, Pasukonis, Ba, and Lillicrap]{hafner2023dreamerv3}
Danijar Hafner, Jurgis Pasukonis, Jimmy Ba, and Timothy Lillicrap.
\newblock Mastering diverse domains through world models.
\newblock \emph{arXiv preprint arXiv:2301.04104}, 2023.

\bibitem[Hansen et~al.(2022)Hansen, Su, and Wang]{Hansen2022tdmpc}
Nicklas Hansen, Hao Su, and Xiaolong Wang.
\newblock Temporal difference learning for model predictive control.
\newblock In \emph{Proceedings of the 39th International Conference on Machine Learning}, Proceedings of Machine Learning Research. PMLR, 2022.
\newblock URL \url{https://proceedings.mlr.press/v162/hansen22a.html}.

\bibitem[Hansen et~al.(2024)Hansen, Su, and Wang]{hansen2024tdmpc2}
Nicklas Hansen, Hao Su, and Xiaolong Wang.
\newblock {TD}-{MPC}2: Scalable, robust world models for continuous control.
\newblock In \emph{The Twelfth International Conference on Learning Representations}, 2024.
\newblock URL \url{https://openreview.net/forum?id=Oxh5CstDJU}.

\bibitem[Hendrycks \& Gimpel(2016)Hendrycks and Gimpel]{hendrycks2016gelu}
Dan Hendrycks and Kevin Gimpel.
\newblock Gaussian error linear units (gelus).
\newblock \emph{arXiv preprint arXiv:1606.08415}, 2016.

\bibitem[Hernandez-Leal et~al.(2020)Hernandez-Leal, Kartal, and Taylor]{Hernandez20masurvey}
Pablo Hernandez-Leal, Bilal Kartal, and Matthew~E. Taylor.
\newblock A very condensed survey and critique of multiagent deep reinforcement learning.
\newblock In \emph{Proceedings of the 19th International Conference on Autonomous Agents and MultiAgent Systems}, 2020.

\bibitem[Iqbal \& Sha(2019)Iqbal and Sha]{iqbal19maac}
Shariq Iqbal and Fei Sha.
\newblock Actor-attention-critic for multi-agent reinforcement learning.
\newblock In \emph{Proceedings of the 36th International Conference on Machine Learning}, Proceedings of Machine Learning Research. PMLR, 2019.
\newblock URL \url{http://proceedings.mlr.press/v97/iqbal19a.html}.

\bibitem[Jaegle et~al.(2021)Jaegle, Gimeno, Brock, Vinyals, Zisserman, and Carreira]{jaegle2021perceiver}
Andrew Jaegle, Felix Gimeno, Andy Brock, Oriol Vinyals, Andrew Zisserman, and Joao Carreira.
\newblock Perceiver: General perception with iterative attention.
\newblock In \emph{International conference on machine learning}, 2021.

\bibitem[Janner et~al.(2019)Janner, Fu, Zhang, and Levine]{janner2019mbpo}
Michael Janner, Justin Fu, Marvin Zhang, and Sergey Levine.
\newblock When to trust your model: Model-based policy optimization.
\newblock In \emph{Advances in Neural Information Processing Systems}, 2019.

\bibitem[Janner et~al.(2021)Janner, Li, and Levine]{janner2021tt}
Michael Janner, Qiyang Li, and Sergey Levine.
\newblock Offline reinforcement learning as one big sequence modeling problem.
\newblock In \emph{Advances in Neural Information Processing Systems}, 2021.

\bibitem[Karpathy(2020)]{Karpathy2020mingpt}
Andrej Karpathy.
\newblock mingpt: A minimal pytorch re-implementation of the openai gpt (generative pretrained transformer) training.
\newblock \url{https://github.com/karpathy/minGPT}, 2020.

\bibitem[Kingma \& Ba(2015)Kingma and Ba]{kingma14adam}
Diederik~P. Kingma and Jimmy Ba.
\newblock Adam: {A} method for stochastic optimization.
\newblock In Yoshua Bengio and Yann LeCun (eds.), \emph{3rd International Conference on Learning Representations}, 2015.
\newblock URL \url{http://arxiv.org/abs/1412.6980}.

\bibitem[Kingma \& Welling(2014)Kingma and Welling]{kingma2022vae}
Diederik~P. Kingma and Max Welling.
\newblock {Auto-Encoding Variational Bayes}.
\newblock In \emph{2nd International Conference on Learning Representations}, 2014.

\bibitem[Kuba et~al.(2021)Kuba, Chen, Wen, Wen, Sun, Wang, and Yang]{kuba2021trust}
Jakub~Grudzien Kuba, Ruiqing Chen, Muning Wen, Ying Wen, Fanglei Sun, Jun Wang, and Yaodong Yang.
\newblock Trust region policy optimisation in multi-agent reinforcement learning.
\newblock In \emph{International Conference on Learning Representations}, 2021.

\bibitem[Liu et~al.(2024)Liu, Ye, Ma, Yang, Liang, and Zhang]{liu2024mazero}
Qihan Liu, Jianing Ye, Xiaoteng Ma, Jun Yang, Bin Liang, and Chongjie Zhang.
\newblock Efficient multi-agent reinforcement learning by planning.
\newblock In \emph{The Twelfth International Conference on Learning Representations}, 2024.
\newblock URL \url{https://openreview.net/forum?id=CpnKq3UJwp}.

\bibitem[Liu et~al.(2020)Liu, Wang, Hu, Hao, Chen, and Gao]{liu20maga}
Yong Liu, Weixun Wang, Yujing Hu, Jianye Hao, Xingguo Chen, and Yang Gao.
\newblock Multi-agent game abstraction via graph attention neural network.
\newblock In \emph{The Thirty-Fourth {AAAI} Conference on Artificial Intelligence}, 2020.
\newblock \doi{10.1609/AAAI.V34I05.6211}.
\newblock URL \url{https://doi.org/10.1609/aaai.v34i05.6211}.

\bibitem[Lowe et~al.(2017)Lowe, Wu, Tamar, Harb, Abbeel, and Mordatch]{lowe17maac}
Ryan Lowe, Yi~Wu, Aviv Tamar, Jean Harb, Pieter Abbeel, and Igor Mordatch.
\newblock Multi-agent actor-critic for mixed cooperative-competitive environments.
\newblock In \emph{Proceedings of the 31st International Conference on Neural Information Processing Systems}, 2017.

\bibitem[Mahajan et~al.(2021)Mahajan, Samvelyan, Mao, Makoviychuk, Garg, Kossaifi, Whiteson, Zhu, and Anandkumar]{mahajan21tesseract}
Anuj Mahajan, Mikayel Samvelyan, Lei Mao, Viktor Makoviychuk, Animesh Garg, Jean Kossaifi, Shimon Whiteson, Yuke Zhu, and Animashree Anandkumar.
\newblock Tesseract: Tensorised actors for multi-agent reinforcement learning.
\newblock In \emph{Proceedings of the 38th International Conference on Machine Learning}, Proceedings of Machine Learning Research. PMLR, 2021.
\newblock URL \url{https://proceedings.mlr.press/v139/mahajan21a.html}.

\bibitem[Micheli et~al.(2023)Micheli, Alonso, and Fleuret]{alonso2023iris}
Vincent Micheli, Eloi Alonso, and Fran{\c{c}}ois Fleuret.
\newblock Transformers are sample-efficient world models.
\newblock In \emph{The Eleventh International Conference on Learning Representations}, 2023.
\newblock URL \url{https://openreview.net/forum?id=vhFu1Acb0xb}.

\bibitem[Nguyen et~al.(2020)Nguyen, Nguyen, and Nahavandi]{nguyen20marl_review}
Thanh~Thi Nguyen, Ngoc~Duy Nguyen, and Saeid Nahavandi.
\newblock Deep reinforcement learning for multiagent systems: A review of challenges, solutions, and applications.
\newblock \emph{IEEE Transactions on Cybernetics}, 50:\penalty0 3826--3839, 2020.
\newblock \doi{10.1109/TCYB.2020.2977374}.

\bibitem[Oliehoek et~al.(2016)Oliehoek, Amato, et~al.]{Oliehoek16dec_pomdp}
Frans~A Oliehoek, Christopher Amato, et~al.
\newblock \emph{\rm A concise introduction to decentralized POMDPs}, volume~1.
\newblock \emph{Springer}, 2016.

\bibitem[Peng et~al.(2021)Peng, Rashid, de~Witt, Kamienny, Torr, Boehmer, and Whiteson]{peng2021facmac}
Bei Peng, Tabish Rashid, Christian~Schroeder de~Witt, Pierre-Alexandre Kamienny, Philip Torr, Wendelin Boehmer, and Shimon Whiteson.
\newblock {FACMAC}: Factored multi-agent centralised policy gradients.
\newblock In \emph{Advances in Neural Information Processing Systems}, 2021.
\newblock URL \url{https://openreview.net/forum?id=WxH774N0mEu}.

\bibitem[Radford et~al.(2019)Radford, Wu, Child, Luan, Amodei, Sutskever, et~al.]{radford2019gpt2}
Alec Radford, Jeffrey Wu, Rewon Child, David Luan, Dario Amodei, Ilya Sutskever, et~al.
\newblock Language models are unsupervised multitask learners.
\newblock \emph{OpenAI blog}, 2019.

\bibitem[Rashid et~al.(2018)Rashid, Samvelyan, Schroeder, Farquhar, Foerster, and Whiteson]{rashid18qmix}
Tabish Rashid, Mikayel Samvelyan, Christian Schroeder, Gregory Farquhar, Jakob Foerster, and Shimon Whiteson.
\newblock {QMIX}: Monotonic value function factorisation for deep multi-agent reinforcement learning.
\newblock In \emph{Proceedings of the 35th International Conference on Machine Learning}, Proceedings of Machine Learning Research. PMLR, 2018.
\newblock URL \url{https://proceedings.mlr.press/v80/rashid18a.html}.

\bibitem[Robine et~al.(2023)Robine, H{\"o}ftmann, Uelwer, and Harmeling]{robine2023twm}
Jan Robine, Marc H{\"o}ftmann, Tobias Uelwer, and Stefan Harmeling.
\newblock Transformer-based world models are happy with 100k interactions.
\newblock In \emph{The Eleventh International Conference on Learning Representations}, 2023.
\newblock URL \url{https://openreview.net/forum?id=TdBaDGCpjly}.

\bibitem[Ryu et~al.(2020)Ryu, Shin, and Park]{Ryu2019MultiAgentAW}
Heechang Ryu, Hayong Shin, and Jinkyoo Park.
\newblock Multi-agent actor-critic with hierarchical graph attention network.
\newblock In \emph{Proceedings of the AAAI Conference on Artificial Intelligence}, 2020.

\bibitem[Samvelyan et~al.(2019)Samvelyan, Rashid, Schroeder~de Witt, Farquhar, Nardelli, Rudner, Hung, Torr, Foerster, and Whiteson]{samvelyan19smac}
Mikayel Samvelyan, Tabish Rashid, Christian Schroeder~de Witt, Gregory Farquhar, Nantas Nardelli, Tim G.~J. Rudner, Chia-Man Hung, Philip H.~S. Torr, Jakob Foerster, and Shimon Whiteson.
\newblock The starcraft multi-agent challenge.
\newblock In \emph{Proceedings of the 18th International Conference on Autonomous Agents and MultiAgent Systems}, 2019.

\bibitem[Schrittwieser et~al.(2020)Schrittwieser, Antonoglou, Hubert, Simonyan, Sifre, Schmitt, Guez, Lockhart, Hassabis, Graepel, Lillicrap, and Silver]{Schrittwieser2019Muzero}
Julian Schrittwieser, Ioannis Antonoglou, Thomas Hubert, Karen Simonyan, L.~Sifre, Simon Schmitt, Arthur Guez, Edward Lockhart, Demis Hassabis, Thore Graepel, Timothy~P. Lillicrap, and David Silver.
\newblock Mastering atari, go, chess and shogi by planning with a learned model.
\newblock \emph{Nature}, 588:\penalty0 604 -- 609, 2020.

\bibitem[Schulman et~al.(2017)Schulman, Wolski, Dhariwal, Radford, and Klimov]{Schulman2017ppo}
John Schulman, Filip Wolski, Prafulla Dhariwal, Alec Radford, and Oleg Klimov.
\newblock Proximal policy optimization algorithms.
\newblock \emph{arXiv preprint arXiv:1707.06347}, 2017.

\bibitem[Son et~al.(2019)Son, Kim, Kang, Hostallero, and Yi]{son19qtran}
Kyunghwan Son, Daewoo Kim, Wan~Ju Kang, David~Earl Hostallero, and Yung Yi.
\newblock {QTRAN}: Learning to factorize with transformation for cooperative multi-agent reinforcement learning.
\newblock In \emph{Proceedings of the 36th International Conference on Machine Learning}, Proceedings of Machine Learning Research. PMLR, 2019.
\newblock URL \url{https://proceedings.mlr.press/v97/son19a.html}.

\bibitem[Sunehag et~al.(2018)Sunehag, Lever, Gruslys, Czarnecki, Zambaldi, Jaderberg, Lanctot, Sonnerat, Leibo, Tuyls, and Graepel]{Sunehag2017vdn}
Peter Sunehag, Guy Lever, Audrunas Gruslys, Wojciech~Marian Czarnecki, Vinicius Zambaldi, Max Jaderberg, Marc Lanctot, Nicolas Sonnerat, Joel~Z. Leibo, Karl Tuyls, and Thore Graepel.
\newblock Value-decomposition networks for cooperative multi-agent learning based on team reward.
\newblock In \emph{Proceedings of the 17th International Conference on Autonomous Agents and MultiAgent Systems}, 2018.

\bibitem[Sutton(1991)]{Sutton1991Dyna}
Richard~S Sutton.
\newblock Dyna, an integrated architecture for learning, planning, and reacting.
\newblock \emph{ACM Sigart Bulletin}, 2\penalty0 (4):\penalty0 160--163, 1991.

\bibitem[Toledo \& Prorok(2024)Toledo and Prorok]{toledo2024codreamer}
Edan Toledo and Amanda Prorok.
\newblock Codreamer: Communication-based decentralised world models.
\newblock \emph{arXiv preprint arXiv:2406.13600}, 2024.

\bibitem[van~den Oord et~al.(2017)van~den Oord, Vinyals, and Kavukcuoglu]{oord17vqvae}
Aaron van~den Oord, Oriol Vinyals, and Koray Kavukcuoglu.
\newblock Neural discrete representation learning.
\newblock In \emph{Advances in Neural Information Processing Systems}, 2017.
\newblock ISBN 9781510860964.

\bibitem[Vaswani et~al.(2017)Vaswani, Shazeer, Parmar, Uszkoreit, Jones, Gomez, Kaiser, and Polosukhin]{vaswani17transformer}
Ashish Vaswani, Noam Shazeer, Niki Parmar, Jakob Uszkoreit, Llion Jones, Aidan~N Gomez, {\L}ukasz Kaiser, and Illia Polosukhin.
\newblock Attention is all you need.
\newblock In \emph{Advances in Neural Information Processing Systems}, 2017.
\newblock URL \url{https://proceedings.neurips.cc/paper_files/paper/2017/file/3f5ee243547dee91fbd053c1c4a845aa-Paper.pdf}.

\bibitem[Wang et~al.(2021)Wang, Ren, Liu, Yu, and Zhang]{wang2021qplex}
Jianhao Wang, Zhizhou Ren, Terry Liu, Yang Yu, and Chongjie Zhang.
\newblock {QPLEX}: Duplex dueling multi-agent q-learning.
\newblock In \emph{International Conference on Learning Representations}, 2021.
\newblock URL \url{https://openreview.net/forum?id=Rcmk0xxIQV}.

\bibitem[Willemsen et~al.(2021)Willemsen, Coppola, and de~Croon]{Willemsen21mambpo}
Dani{\"e}l Willemsen, Mario Coppola, and Guido~CHE de~Croon.
\newblock Mambpo: Sample-efficient multi-robot reinforcement learning using learned world models.
\newblock In \emph{2021 IEEE/RSJ International Conference on Intelligent Robots and Systems (IROS)}. IEEE, 2021.

\bibitem[Xu et~al.(2022)Xu, Li, Zhang, Zhan, Baiia, and Fan]{xu2022mbvd}
Zhiwei Xu, Dapeng Li, Bin Zhang, Yuan Zhan, Yunpeng Baiia, and Guoliang Fan.
\newblock Mingling foresight with imagination: Model-based cooperative multi-agent reinforcement learning.
\newblock In \emph{Advances in Neural Information Processing Systems}, 2022.
\newblock URL \url{https://openreview.net/forum?id=flBYpZkW6ST}.

\bibitem[Yu et~al.(2022)Yu, Velu, Vinitsky, Gao, Wang, Bayen, and Wu]{yu2022mappo}
Chao Yu, Akash Velu, Eugene Vinitsky, Jiaxuan Gao, Yu~Wang, Alexandre Bayen, and Yi~Wu.
\newblock The surprising effectiveness of {PPO} in cooperative multi-agent games.
\newblock In \emph{Thirty-sixth Conference on Neural Information Processing Systems Datasets and Benchmarks Track}, 2022.

\bibitem[Zhang et~al.(2024{\natexlab{a}})Zhang, Bai, Hu, Wang, and Li]{zhang2024provably}
Qiaosheng Zhang, Chenjia Bai, Shuyue Hu, Zhen Wang, and Xuelong Li.
\newblock Provably efficient information-directed sampling algorithms for multi-agent reinforcement learning.
\newblock \emph{arXiv preprint arXiv:2404.19292}, 2024{\natexlab{a}}.

\bibitem[Zhang et~al.(2024{\natexlab{b}})Zhang, Yang, Bai, Wu, Li, Li, and Wang]{zhang2024read}
Yang Zhang, Shixin Yang, Chenjia Bai, Fei Wu, Xiu Li, Xuelong Li, and Zhen Wang.
\newblock Towards efficient llm grounding for embodied multi-agent collaboration.
\newblock \emph{arXiv preprint arXiv:2405.14314}, 2024{\natexlab{b}}.

\bibitem[Łukasz Kaiser et~al.(2020)Łukasz Kaiser, Babaeizadeh, Miłos, Osiński, Campbell, Czechowski, Erhan, Finn, Kozakowski, Levine, Mohiuddin, Sepassi, Tucker, and Michalewski]{Kaiser2020SimPLe}
Łukasz Kaiser, Mohammad Babaeizadeh, Piotr Miłos, Błażej Osiński, Roy~H Campbell, Konrad Czechowski, Dumitru Erhan, Chelsea Finn, Piotr Kozakowski, Sergey Levine, Afroz Mohiuddin, Ryan Sepassi, George Tucker, and Henryk Michalewski.
\newblock Model based reinforcement learning for atari.
\newblock In \emph{International Conference on Learning Representations}, 2020.
\newblock URL \url{https://openreview.net/forum?id=S1xCPJHtDB}.

\end{thebibliography}
\bibliographystyle{tmlr}
\newpage
\appendix
\section{World Models Details and Hyperparameters}\label{appendix:model_details}

\subsection{Observation Tokenizer}
Our tokenizer for local observation discretization is based on the implementation\footnote{Code can be found in \url{https://github.com/lucidrains/vector-quantize-pytorch}} of a vanilla VQ-VAE \citep{oord17vqvae}.
Faced with continuous non-vision observation, we build the encoder and decoder as Multi-Layer Perceptrons (MLPs). The decoder is designed with the same hyperparameters as the ones of the encoder.
The hyperparameters are listed as Table~\ref{table:vqvae_hyper}. During the phase of collecting experience from the external environment, each agent takes the reconstructed observations processed by the VQ-VAE as input instead to avoid the distribution shift between policy learning and policy execution.

For training this vanilla VQ-VAE, we use a straight-through estimator to enable gradient backpropagation through the non-differentiable quantization operation in the quantization of VQ-VAE. The loss function for learning the autoencoder is as follows:
\begin{equation}\label{equ:loss_vqvae}
    \mathcal{L}_{\rm VQ-VAE}(E, D, \mathcal{Z}) = \mathbb{E}_{i \sim \mathcal{N}} \mathbb{E}_{o^i} \left[ \| o^i - \hat{o}^i \|^{2} + \| {\rm sg}[E(o^i)] - z_{q}^{i} \|^{2} + \beta \| {\rm sg}[z_{q}^{i}] - E(o^i) \|^{2} \right]
\end{equation}
where $\mathcal{N} = \{1, 2, ..., n\}$ denotes the set of agents, ${\rm sg}[\cdot]$ denotes the stop-gradient operation and $\beta$ is the coefficient of the commitment loss $\| {\rm sg}[z_{q}^{i}] - E(o^i) \|^{2}$.
In practice, we found the codebook $\mathcal{Z}$ can suffer from codebook collapse when learning from scratch. Thus, we adopt the Exponential Moving Averages (EMA) \citep{oord17vqvae} technique to alleviate this problem.

\begin{table}[h]
  \caption{VQVAE hyperparameters.}
  \label{table:vqvae_hyper}
  \centering
  \begin{tabular}{ll}
    \toprule
    {\bf Hyperparameter}   &  {\bf Value}                   \\
    \midrule
    \textbf{\underline{Encoder\&Decoder}}   &               \\
    Layers           &  3                                   \\
    Hidden size      &  512                                 \\
    Activation       &  GELU\citep{hendrycks2016gelu}       \\
    \textbf{\underline{Codebook}}   &                       \\
    Codebook size ($N$)     &  512                          \\
    Tokens per observation ($K$)    &  16                   \\
    Code dimension          &  128                          \\
    Coef. of commitment loss ($\beta$)   &  10.0            \\
    \bottomrule
  \end{tabular}
\end{table}

\subsection{Transformer}\label{appendix:trans_detail}
The shared Transformer serving as the local dynamics model is based on the implementation of minGPT \citep{Karpathy2020mingpt}. Given a fixed imagination horizon $H$, it first takes a token sequence of length $H(K + 1)$ composed of observation tokens and action tokens, and embeds it into a $H(K + 1) \times D$ tensor via separate embedding tables for observations and actions. Then, the aggregated feature tensor, returned by the agent-wise aggregation module, is inserted after the action embedding tensor at every timestep, forming a final embedding tensor of shape $H(K + 2) \times D$. This tensor is forwarded through fixed Transformer blocks. Here, we adopt GPT2-like blocks \citep{radford2019gpt2} as the basic blocks.
The hyperparameters are listed as Table~\ref{table:trans_hyper}. To enable training across all environments on a single NVIDIA RTX 3090 GPU, we adapt imagination horizon $H$ based on the number of agents.
\begin{table}[h]
  \caption{Transformer hyperparameters.}
  \label{table:trans_hyper}
  \centering
  \begin{tabular}{ll}
    \toprule
    {\bf Hyperparameter}   &  {\bf Value}                   \\
    \midrule
    Imagination horizon ($H$)     &     \{15, 8, 5\}              \\
    Embedding dimension           &     256                 \\
    Layers                        &     10                  \\
    Attention heads               &     4                   \\
    Weight decay                  &     0.01                \\
    Embedding dropout             &     0.1                 \\
    Attention dropout             &     0.1                 \\
    Residual dropout              &     0.1                 \\
    \bottomrule
  \end{tabular}
\end{table}

\subsection{Perceiver}
The Perceiver \citep{jaegle2021perceiver} is based on the open-source implementation\footnote{Code can be found in \url{https://github.com/lucidrains/perceiver-pytorch}}.
By aligning the length of the latent querying array with the number of agents $n$, we obtain the intrinsic global representation feature corresponding to each individual agent.
We further dive into the process of agent-wise representation aggregation: (\romannumeral1) the embedding tensor of shape $(K + 1) \times D$ at each timestep, mentioned in Appendix~\ref{appendix:trans_detail}, is concatenated with others from all agents, thereby getting a $n(K + 1) \times D$ sequence for the joint observation-action pair at the current timestep; (\romannumeral2) through the cross-attention mechanism with the latent querying array, the original sequence is compressed from length $n(K + 1)$ to $n$; (\romannumeral3) the compressed sequence is then forwarded through a standard transformer with bidirectional attention inside the Perceiver. The hyperparameters are listed as Table~\ref{table:perceiver_hyper}.
\begin{table}
  \caption{Perceiver hyperparameters.}
  \label{table:perceiver_hyper}
  \centering
  \begin{tabular}{ll}
    \toprule
    {\bf Hyperparameter}   &  {\bf Value}                   \\
    \midrule
    Length of latent querying     &    $n$ (number of agents)       \\
    Cross attention heads         &    8                            \\
    Inner Transformer layers      &    2                            \\
    Transformer attention heads   &    8                            \\
    Dimension per attention head  &     64                          \\
    Embedding dropout             &     0.1                         \\
    Attention dropout             &     0.1                         \\
    Residual dropout              &     0.1                         \\
    \bottomrule
  \end{tabular}
\end{table}


\section{Behaviour Learning Details}\label{appendix:behaviour_learning}
In MARIE, we use MAPPO-like \citep{yu2022mappo} actor and critic, where the actor and critic should have been 3-layer MLPs. However, unlike other CTDE model-free approaches, whose critic takes additional global oracle states from the environment in the training phase, our world model hardly provides related predictions in the imagined trajectories. To alleviate this issue, we augment the critic with an attention mechanism and provide it all reconstructed observations $\boldsymbol{\hat{o}}_{t}$ of all agents. 
Therefore, the actor $\psi$ remains a 3-layer MLP with ReLU activation, while the critic $\xi$ is enhanced with an extra layer of self-attention, built on top of the original 3-layer MLP, i.e., we overwrite the critic $V_{\xi}^{i}(\boldsymbol{\hat{o}}_{t}) \simeq \mathbb{E}_{\pi_{\psi}^{i}} (\sum\nolimits_{l \geq t} \gamma^{l - t}\hat{r}^{i}_{l})$ for agent $i$.
Similar to off-the-shelf CTDE model-free approaches, we adopt parameter sharing across agents.
\paragraph{Critic loss function}
We utilize $\lambda$-return in Dreamer \citep{hafner2019dreamer}, which employs an exponentially-weighted average of different $k$-steps TD targets to balance bias and variance as the regression target for the critic.
Given an imagined trajectory $\{ \hat{o}_{\tau}^{i}, a_{\tau}^{i}, \hat{r}_{\tau}^{i}, \hat{\gamma}_{\tau}^{i} \}^{H}_{t = 1}$ for agent $i$, $\lambda$-return is calculated recursively as,
\begin{equation}
    V_{\lambda}^{i}(\boldsymbol{\hat{o}}_{t}) = \left\{\begin{array}{lcl}
    \hat{r}_{t}^{i} + \hat{\gamma}_{t}^{i} \left[ (1 - \lambda) V_{\xi}^{i}(\boldsymbol{\hat{o}}_{t}) + \lambda V_{\lambda}^{i}(\boldsymbol{\hat{o}}_{t+1}) \right]  &  \mbox{if}  & t < H \\
    V_{\xi}^{i}(\boldsymbol{\hat{o}}_{t})     &  \mbox{if}  & t = H
    \end{array}\right.
\end{equation}
The objective of the critic $\xi$ is to minimize the mean squared difference $\mathcal{L}_{\xi}^{i}$ with $\lambda$-returns over imagined trajectories for each agent $i$, as
\begin{equation}
\label{equ:critic_loss}
    \mathcal{L}_{\xi}^{i} = \mathbb{E}_{\pi_{\psi}^{i}} \left[ \sum\nolimits_{t = 1}^{H - 1} \left( V_{\xi}^{i}(\boldsymbol{\hat{o}}_{t}) - {\rm sg}(V_{\lambda}^{i}(\boldsymbol{\hat{o}}_{t})) \right)^{2} \right]
\end{equation}
where ${\rm sg}(\cdot)$ denotes the stop-gradient operation. We optimize the critic loss with respect to the critic parameters $\xi$ using the Adam optimizer.

\paragraph{Actor loss function}
The objective for the action model $\pi_{\psi}(\cdot | \hat{o}_{t}^{i})$ is to output actions that maximize the prediction of long-term future rewards made by the critic. To incorporate intermediate rewards more directly, we train the actor to maximize the same $\lambda$-return that was computed for training the critic. In terms of the non-stationarity issue in multi-agent scenarios, we adopt PPO updates, which introduce important sampling for actor learning. The actor loss function for agent $i$ is:
\begin{align}
\label{equ:actor_loss}
    \mathcal{L}_{\psi}^{i} = -\mathbb{E}_{p_{\phi}, \pi_{\psi_{\rm old}}^{i}} \Big[ \sum_{t=0}^{H - 1} \mathop{\min} \Big(r_{t}^{i}(\psi) A_{t}^{i}, {\rm clip}(r_{t}^{i}(\psi), 1 - \epsilon, 1 + \epsilon) A_{t}^{i}\Big) + \eta \mathcal{H}(\pi_{\psi}^{i}(\cdot | \hat{o}_{t}^{i})) \Big]
\end{align}
where $r_{t}^{i}(\psi) = \pi_{\psi}^{i} / \pi_{\psi_{\rm old}}^{i}$ is the policy ratio and $A_{t}^{i} = {\rm sg}(V_{\lambda}^{i}(\boldsymbol{\hat{o}}_{t}) - V_{\xi}^{i} (\boldsymbol{\hat{o}}_{t}))$ is the advantage. We optimize the actor loss with respect to the actor parameters $\psi$ using the Adam optimizer.
In the discount prediction of MARIE, we set its learning target $\gamma$ to be 0.99. Overall hyperparameters are shown in Table~\ref{table:rl_hyper}.
\begin{table}
  \caption{Behaviour learning hyperparameters.}
  \label{table:rl_hyper}
  \centering
  \begin{tabular}{ll}
    \toprule
    {\bf Hyperparameter}   &  {\bf Value}                   \\
    \midrule
    Imagination Horizon ($H$)               &    \{15, 8, 5\}             \\
    Predicted discount label $\gamma$       &    0.99                  \\
    $\lambda$                               &    0.95               \\
    $\eta$                                  &    0.001              \\
    Clipping parameter $\epsilon$           &    0.2                \\
    \bottomrule
  \end{tabular}
\end{table}

\section{Extended Analysis on Attention Patterns}\label{appendix:attn_vis}
\begin{figure*}[t]
\vskip -0.05in
\begin{center}
\centerline{
\includegraphics[width=1.0\columnwidth]{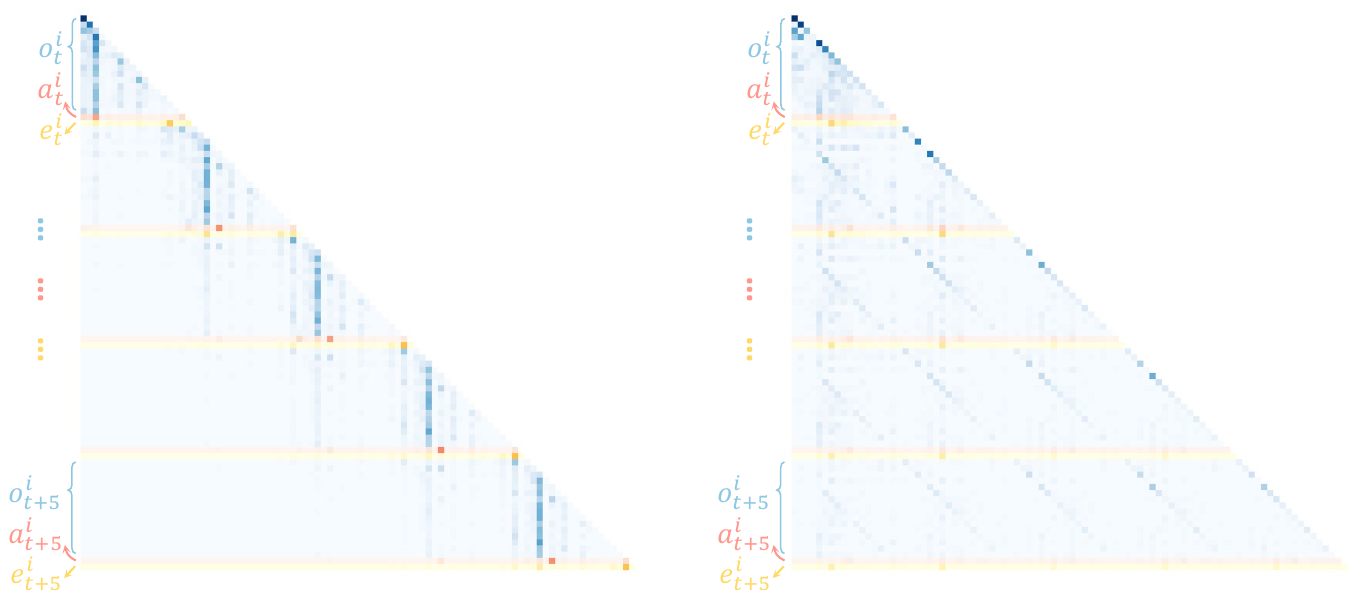}}
\vspace{-0.5em}
\caption{{\bf Attention patterns in the Transformer.}
We observe two distinct types of attention weights during the prediction of local dynamics. In the first one (\textbf{\emph{left}}), the next observation prediction is primarily dependent on the last transition, which means the world model has learned the Markov property corresponding to Dec-POMDPs.
The second type (\textbf{\emph{right}}) exhibits a regularly striated pattern, where the next observation prediction hinges overwhelmingly on the same dimension of multiple previous timesteps.
The above attention weights are produced by a sixth-layer and ninth-layer attention head during imaginations on the \emph{3s\_vs\_5z} scenario.
}
\label{fig:attn_vis}
\end{center}
\vspace{-2em}
\end{figure*}

\begin{figure*}[t]
\vskip -0.05in
\begin{center}
\centerline{
\includegraphics[width=1.0\columnwidth]{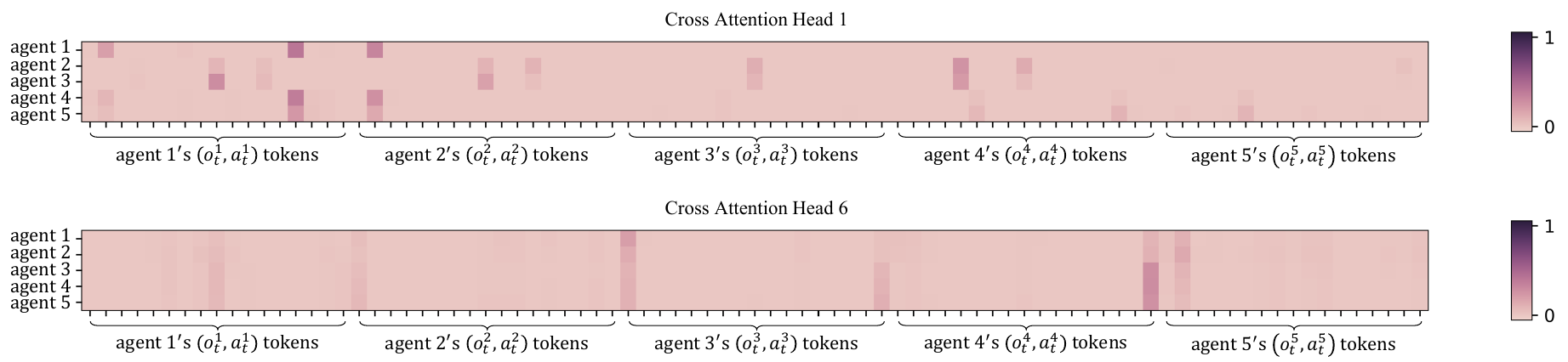}}
\vspace{-0.5em}
\caption{{\bf Cross attention patterns in the Perceiver.}
We observe the \emph{individuality} and \emph{commonality} in the agent-wise aggregation.
The top part of the figure represents the \emph{individuality}, where agents adjust their attentions over the whole joint 
token sequence at timestep $t$ flexibly according to their own needs.
In contrast, the bottom exhibits the \emph{commonality}, where every agent's attention over the joint token sequence is emphasized in the similar positions of the sequence.
The cross attention weights mentioned above are produced by the first and sixth head of the cross attention within the Perceiver, during the agent-wise aggregation on the \emph{2s3z} scenario.}
\label{fig:cross_attn_vis}
\end{center}
\vspace{-2em}
\end{figure*}

To provide qualitative analysis of our world model, we select typical scenarios -- \emph{3s\_vs\_5z} where our method achieves the most significant improvement compared to other baselines for visualizing attention maps inside the Transformer. 
For the sake of simple and clear visualization, we set the imagination horizon $H$ as 5.
In terms of cross-attention maps in the aggregation module, we select a scenario \emph{2s3z} including 5 agents for visualization.
Visualization results are depicted as Fig.~\ref{fig:attn_vis} and Fig.~\ref{fig:cross_attn_vis}.

The prediction of local dynamics entails two distinct attention patterns. The left one in Fig.~\ref{fig:attn_vis} can be interpreted as a Markovian pattern, in which the observation prediction lays its focus on the previous transition.
In contrast, the right one is regularly striated, with the model attending to specific tokens in multiple prior observations.
In terms of the agent-wise aggregation, we also identify two distinct patterns: \emph{individuality} and \emph{commonality}. The top one in Fig.~\ref{fig:cross_attn_vis} illustrates that each agent flexibly attends to different tokens according to their specific needs. In contrast, the bottom one exhibits consistent attention allocation across all agents, with attention highlighted in nearly identical positions.
The diverse patterns in the Transformer and Perceiver may be the key to accurate and consistent imagination.

\section{Baseline Implementation Details}\label{app:baseline_hyper}
\textbf{MAMBA} \citep{Egorov22mamba} is evaluated based on the open-source implementation: \url{https://github.com/jbr-ai-labs/mamba} with the hyperparameters in Table~\ref{table:mamba_hyper}.
\begin{table}
  \caption{Hyperparameters for MAMBA in SMAC environments.}
  \label{table:mamba_hyper}
  \centering
  \begin{tabular}{ll}
    \toprule
    {\bf Hyperparameter}   &  {\bf Value}                   \\
    \midrule
    Batch size     &    256       \\
    $\lambda$ for $\lambda$-return computation        &    0.95                            \\
    Entropy coefficient      &    0.001                            \\
    Entropy annealing   &    0.99998                            \\
    Number of policy updates  &     4                          \\
    Epochs per policy update             &     5                         \\
    Clipping parameter $\epsilon$           &     0.2                         \\
    Actor Learning rate              &     0.0005                         \\
    Critic Learning rate              &     0.0005                         \\
    Discount factor $\gamma$        &   0.99                   \\
    Model Learning rate             &   0.0002                  \\
    Number of model training epochs       &     60              \\
    Number of imagined rollouts           &     800              \\
    Sequence length                       &     20              \\
    Imagination horizon $H$               &     15              \\
    Buffer size                           &     $2.5\times 10^5$  \\
    Number of categoricals                &     32              \\
    Number of classes                     &     32              \\
    KL balancing entropy weight           &     0.2             \\
    KL balancing cross entropy weight           &     0.8             \\
    Gradient clipping                     &     100             \\
    Collected trajectories between updates          &     1     \\
    Hidden size                           &     256             \\
    \bottomrule
  \end{tabular}
\end{table}

\textbf{MAPPO} \citep{yu2022mappo} is evaluated based on the open-source implementation: \url{https:// github.com/marlbenchmark/on-policy} with the common hyperparameters in Table~\ref{table:mappo_hyper}.
\begin{table}
  \caption{Common hyperparameters for MAPPO in SMAC environments.}
  \label{table:mappo_hyper}
  \centering
  \begin{tabular}{ll}
    \toprule
    {\bf Hyperparameter}   &  {\bf Value}                   \\
    \midrule
    Batch size              &     num envs × buffer length × num agents \\
    Mini batch size         &     batch size / mini-batch      \\
    Recurrent data chunk length     &    10       \\
    GAE $\lambda$         &    0.95                            \\
    Discount factor $\gamma$        &   0.99                   \\
    Value loss        &     huber loss                          \\
    Huber delta             &     10.0                         \\
    Optimizer           &     Adam                         \\
    Optimizer learning rate     &    0.0005                 \\
    Optimizer epsilon             &     $1\times 10^{-5}$                         \\
    Weight decay              &     0.0                         \\
    Gradient clipping                     &     10             \\
    Network initialization             &   orthogonal                  \\
    Use reward normalization       &     True              \\
    Use feature normalization           &     True              \\
    \bottomrule
  \end{tabular}
\end{table}

\textbf{QMIX} \citep{rashid18qmix} is evaluated based on the open-source implementation: \url{https:// github.com/oxwhirl/pymarl} with the hyperparameters in Table~\ref{table:qmix_hyper}.
\begin{table}
  \caption{Hyperparameters for QMIX in SMAC environments.}
  \label{table:qmix_hyper}
  \centering
  \begin{tabular}{ll}
    \toprule
    {\bf Hyperparameter}   &  {\bf Value}                   \\
    \midrule
    Batch size              &     32  \\
    Buffer size         &     5000      \\
    Epsilon in epsilon-greedy     &    $1.0 \rightarrow 0.05$       \\
    Epsilon anneal time              &      50000    \\
    Train interval                   &      1 episode    \\
    Discount factor $\gamma$        &   0.99                   \\
    Optimizer           &     RMSProp                         \\
    RMSProp $\alpha$     &    0.99                 \\
    RMSProp $\epsilon$     &    $10^{-5}$                 \\
    Gradient clipping                     &     10             \\
    \bottomrule
  \end{tabular}
\end{table}

\textbf{QPLEX} \citep{wang2021qplex} is evaluated based on the open-source implementation: \url{https:// github.com/wjh720/QPLEX} with the hyperparameters in Table~\ref{table:qplex_hyper}. Since its implementation is mostly based on the open-source implementation: \href{https:// github.com/oxwhirl/pymarl}{PyMARL} \citep{samvelyan19smac}, its most hyperparameters setting remains the same as the one in QMIX in addition to its own special hyperparameters.
\begin{table}
  \caption{Hyperparameters for QPLEX in SMAC environments.}
  \label{table:qplex_hyper}
  \centering
  \begin{tabular}{ll}
    \toprule
    {\bf Hyperparameter}   &  {\bf Value}                   \\
    \midrule
    Batch size              &     32 \\
    Buffer size         &     5000     \\
    Epsilon in epsilon-greedy     &    $1.0 \rightarrow 0.05$       \\
    Epsilon anneal time              &      50000    \\
    Train interval                   &      1 episode    \\
    Discount factor $\gamma$        &   0.99                   \\
    Optimizer           &     RMSProp                         \\
    RMSProp $\alpha$     &    0.99                 \\
    RMSProp $\epsilon$     &    $10^{-5}$                 \\
    Gradient clipping                     &     10             \\
    Number of layers in HyperNetwork                     &     1             \\
    Number of heads in the attention module                     &     4             \\
    \bottomrule
  \end{tabular}
\end{table}

\textbf{MBVD} \citep{xu2022mbvd} is evaluated based on the implementation in its supplementary material from \url{https://openreview.net/forum?id=flBYpZkW6ST} with the hyperparameters in Table~\ref{table:mbvd_hyper}.
Akin to QPLEX, its implementation is based on the open-source implementation: \href{https:// github.com/oxwhirl/pymarl}{PyMARL}, its most hyperparameters setting remains the same as the one in QMIX in addition to its own special hyperparameters.
\begin{table}
  \caption{Hyperparameters for MBVD in SMAC environments.}
  \label{table:mbvd_hyper}
  \centering
  \begin{tabular}{ll}
    \toprule
    {\bf Hyperparameter}   &  {\bf Value}                   \\
    \midrule
    Batch size              &     32 \\
    Buffer size         &     5000     \\
    Epsilon in epsilon-greedy     &    $1.0 \rightarrow 0.05$       \\
    Epsilon anneal time              &      50000    \\
    Train interval                   &      1 episode    \\
    Discount factor $\gamma$        &   0.99                   \\
    Optimizer           &     RMSProp                         \\
    RMSProp $\alpha$     &    0.99                 \\
    RMSProp $\epsilon$     &    $10^{-5}$                 \\
    Gradient clipping                     &     10             \\
    Number of layers in HyperNetwork                     &     1             \\
    Number of heads in the attention module                     &     4             \\
    Horizon of the imagined rollout                             &     3         \\
    KL balancing $\alpha$                   &       0.3             \\
    Dimension of the latent state $\hat{s}$     &       num agents x 16   \\
    \bottomrule
  \end{tabular}
\end{table}

\section{Additional Experiments}
\subsection{Evaluations on MAMujoco}\label{appendix:mamujoco_exp}
The Multi-Agent MuJoCo (MAMuJoCo) \citep{peng2021facmac} environment is a multi-agent extension of MuJoCo. While the MuJoCo tasks challenge a robot to learn an optimal way of motion, MAMuJoCo models each part of a robot as an independent agent — for example, a leg for a spider or an arm for a swimmer — and requires the agents to collectively perform efficient motion. With the increasing variety of the body parts, MAMujoco can be also considered as a testbed for evaluating the coordination among heterogeneous agents, which poses a big challenge for learning the multi-agent dynamics inside it, especially in a \emph{decentralized} manner.

\begin{figure*}[t!]
\vskip -0.05in
\begin{center}
\centerline{
\includegraphics[width=1.0\columnwidth]{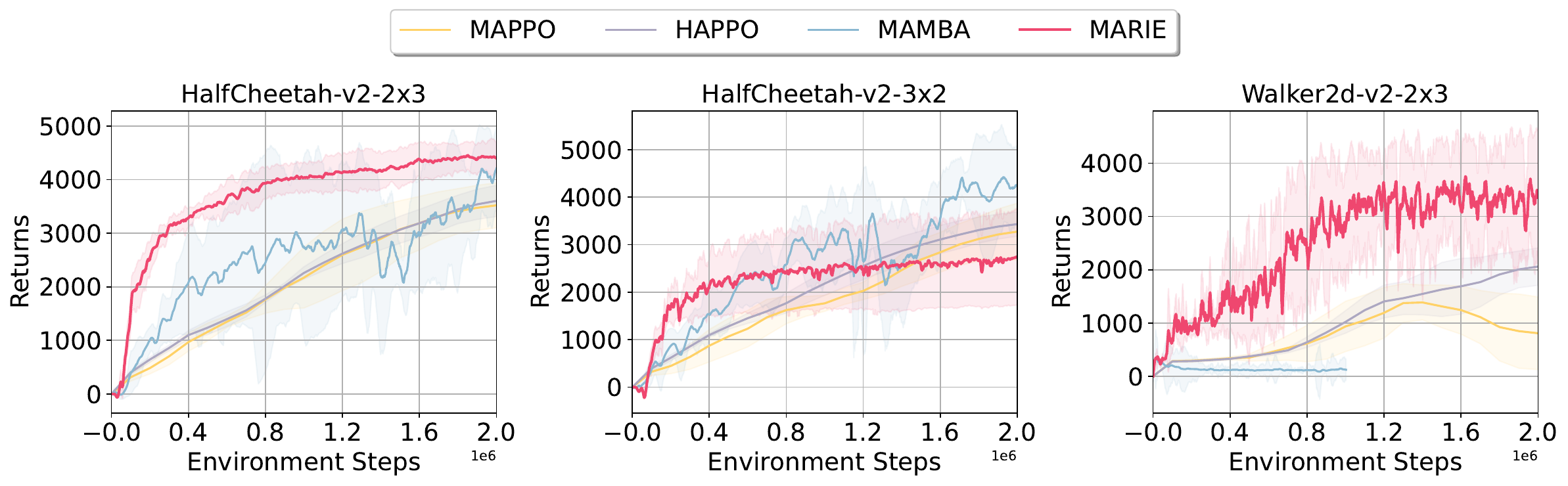}
}
\vspace{-0.5em}
\caption{
Curves of the performance for MARIE, MAMBA, HAPPO and MAPPO in 3 chosen MAMujoco scenarios. Y axis: return; X axis: number of steps taken in the real environment.}
\label{fig:mamujoco_exp_result}
\end{center}
\vspace{-2em}
\end{figure*}

While MAMBA \citep{Egorov22mamba} originally does not take the continuous action space case into consideration, which is a obvious limitation in it, we would like to evaluate MARIE in such case, e.g., MAMujoco, to better demonstrate that our method can be also effective in other multi-agent domains.
In MAMujoco, we discretize the scalar in each dimension of continuous actions into one of $256$ fixed-width bins independently to obtain discrete action tokens for local dynamics learning.
As the behaviour learning in MARIE adopts a MAPPO-like and on-policy manner, we choose two strong on-policy PPO-based baselines -- MAPPO \citep{yu2022mappo} and HAPPO \citep{kuba2021trust}. {Additionally, we also include MAMBA as a model-based baseline for comparison.}
{Since MAMBA\citep{Egorov22mamba} was originally originally designed for domains with discrete action space, significant effort was required to adapt and evaluate it on MAMujoco, which features continuous action space.}
The experiments are conducted in \emph{HalfCheetah-v2-2x3}, \emph{HalfCheetah-v2-3x2} and \emph{Walker2d-v2-2x3}.
The learning curves of the return averaged over 4 seeds are presented as Figure~\ref{fig:mamujoco_exp_result}.
{Notably, MAMBA fails to enhance policy learning in the \emph{Walker2d-v2-2x3} scenario and remains exceptionally time-consuming. Consequently, we report its results only for 1 million environment steps in this scenario.}

As illustrated in Figure~\ref{fig:mamujoco_exp_result}, our MARIE consistently shows superior sample efficiency and achieves the best performance in {2 of 3} scenarios with limited 2M environment steps.
{For the performance difference between MAMBA and MARIE in \emph{HalfCheetah-v2-3x2}, we hypothesize that MAMBA's policy learning benefits significantly from using the internal recurrent features of the world model as inputs in this scenario, while the policy in our method only takes the reconstructed observation as input in order to support fast deployment in the environment without the participation of the world model.}
We attribute the performance gap between MARIE and other two model-free baselines in \emph{HalfCheetah-v2-3x2} to the access to global oracle state in the chosen baselines.
The policy in our algorithm is purely learned from the inner imaginations of the world model where there is only reconstructed local observation. Considering MAMujoco is a multi-heterogeneous-agent benchmark which necessitates a more precise credit assignment during training, it would be much more helpful for policy learning to have access to the true global oracle state than in other benchmarks.
But overall, our MARIE presents a faster convergence rate, implying that our Transformer-based world model can generate accurate imaginations and bring remarkable sample efficiency.

\subsection{Evaluations on SMACv2}\label{appendix:smacv2_exp}
\begin{figure*}[t!]
\vskip -0.05in
\begin{center}
\centerline{
\includegraphics[width=1.0\columnwidth]{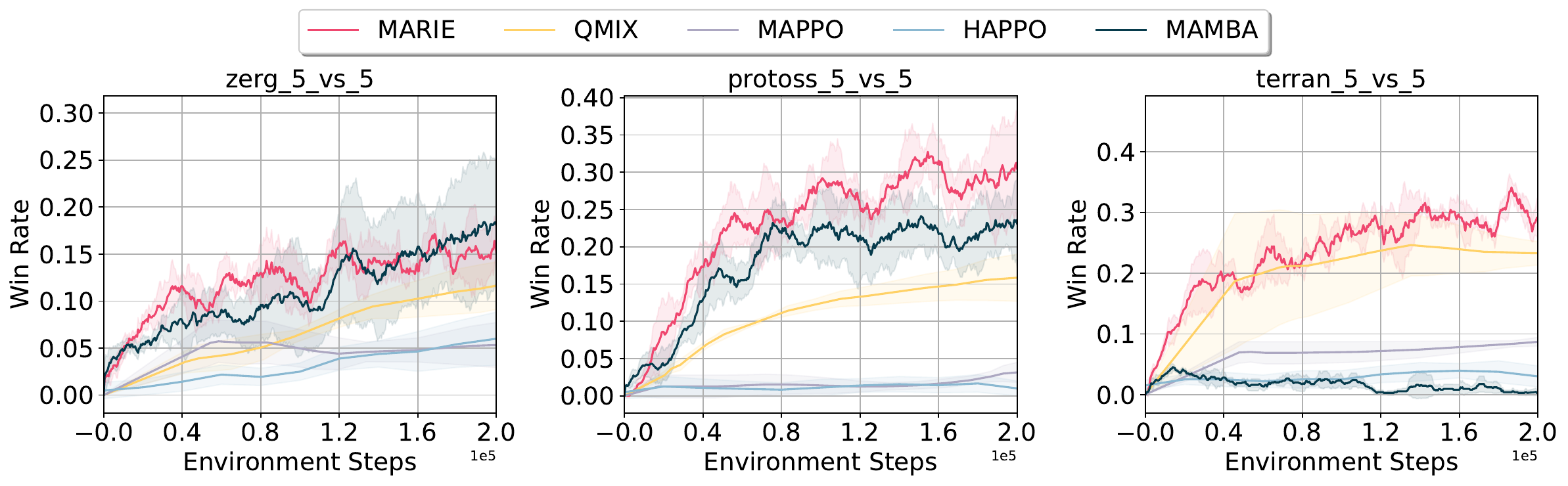}
}
\vspace{-0.5em}
\caption{
{Curves of the performance for MARIE, MAMBA, QMIX, HAPPO and MAPPO in 3 chosen SMACv2 scenarios. Y axis: Win Rate; X axis: number of steps taken in the real environment. While MARIE shows competitive performance to MAMBA on zerg\_5\_vs\_5, MARIE is superior to all other baselines in terms of sample efficiency and final performance in the rest 2 scenarios.}}
\label{fig:smacv2_result}
\end{center}
\vspace{-2em}
\end{figure*}

{Given known serious flaws in SMACv1 (e.g., the tricky open-loop policy issue), we extend our evaluation of MARIE to SMACv2 \citep{ellis2023smacv2}, which introduces more stochasticity and partial observability. In this comparison, we benchmark MARIE against four baselines: MAPPO, HAPPO, QMIX and MAMBA. For each random seed, we adopt the same evaluation protocol as the main experiment on SMACv1. Importantly, the hyperparameters of MARIE remain unchanged, as detailed in \S\ref{appendix:marie_hyper}.}
{Here, we directly use the results of MAPPO and QMIX provided in the official SMACv2 repository\footnote{Results of QMIX and MAPPO are available at \url{https://github.com/oxwhirl/smacv2/tree/main/smacv2/examples/results}.}.}
{Illustrated in Figure~\ref{fig:smacv2_result}, while MARIE shows competitive performance to MAMBA on zerg\_5\_vs\_5, MARIE is superior to all other baselines in terms of sample efficiency and final performance in the rest 2 scenarios.}

\subsection{Comparison with QMIX and QPLEX with different epsilon annealing time}
Considering the potentially inappropriate influence of a large $\epsilon$ annealing time used in the epsilon-greedy algorithm when evaluated in the low data regime evaluation, we run QMIX and QPLEX with a smaller $\epsilon$ annealing time, and compare the performance of them with ours.
The result is reported in Table~\ref{table:small_t_epsilon}.
The reported result shows that the original hyperparameters used in the main experiment, which are also directly referred to \citet{Egorov22mamba} and \citet{liu2024mazero}, are reasonable since the performance of QMIX and QPLEX under the original hyperparameters is superior to the ones with a smaller $\epsilon$ annealing time at most scenarios.
Besides, our MARIE still consistently outperforms QMIX and QPLEX with a smaller $\epsilon$ annealing time.

\begin{table}[htb!]
\centering
\caption{Mean evaluation win rate and standard deviation for QMIX and QPLEX with different epsilon anneal time $t_{\epsilon}$ over 4 random seeds. We bold the values of the maximum.}
\vspace{1em}
\resizebox{\columnwidth}{!}{%
\begin{tabular}{ccccccc}
\toprule
Maps  &  Steps  &  MARIE  &  QMIX ($t_{\epsilon}  = 50000$)   & QMIX ($t_{\epsilon}  = 10000$)   & QPLEX ($t_{\epsilon}  = 50000$)  & QPLEX ($t_{\epsilon}  = 10000$)  \\ \midrule
1c3s5z & \multirow{10}{*}{100K} & \colorbox{mine}{\textbf{85.0}\scr{(9.4)}} & 43.6\scr{(29.2)} & 33.3\scr{(15.0)} & 68.3\scr{(7.4)} & 44.8\scr{(11.0)}  \\
2m\_vs\_1z  & & \colorbox{mine}{\textbf{95.5}\scr{(7.9)}} & 70.3\scr{(14.8)} & 36.1\scr{(28.2)} & 84.8\scr{(10.8)} & 93.2\scr{(4.7)} \\
2s\_vs\_1sc & & \colorbox{mine}{\textbf{96.9}\scr{(7.1)}} & 0.0\scr{(0.0)} & 3.9\scr{(6.7)} & 15.7\scr{(19.5)} & 43.2\scr{(32.4)} \\
2s3z & & \colorbox{mine}{\textbf{80.5}\scr{(9.3)}} & 37.7\scr{(15.5)} & 29.1\scr{(20.3)} & 50.2\scr{(8.4)} & 28.3\scr{(11.5)} \\
3m & & \colorbox{mine}{\textbf{99.5}\scr{(0.4)}} & 54.4\scr{(22.7)} & 63.8\scr{(14.6)} & 88.7\scr{(6.9)} & 85.0\scr{(11.3)} \\
3s\_vs\_3z & & \colorbox{mine}{\textbf{98.9}\scr{(1.5)}} & 0.0\scr{(0.0)} & 0.0\scr{(0.0)} & 0.0\scr{(0.0)} & 0.0\scr{(0.0)} \\ 
3s\_vs\_4z & & \colorbox{mine}{\textbf{73.0}\scr{(6.2)}} & 0.0\scr{(0.0)} & 0.0\scr{(0.0)} & 0.0\scr{(0.0)} & 0.0\scr{(0.0)} \\
8m & & \colorbox{mine}{\textbf{88.0}\scr{(3.9)}} & 69.5\scr{(12.8)} & 68.6\scr{(13.6)} & 83.4\scr{(6.4)} & 79.7\scr{(9.8)} \\
MMM & & \colorbox{mine}{\textbf{87.6}\scr{(3.0)}} & 31.1\scr{(17.3)} & 18.9\scr{(4.3)} & 69.3\scr{(35.1)} & 20.2\scr{(7.7)} \\
so\_many\_baneling & & \colorbox{mine}{\textbf{94.8}\scr{(5.9)}} & 20.0\scr{(8.9)} & 30.7\scr{(18.5)} & 32.2\scr{(6.1)} & 37.7\scr{(9.2)} \\ \midrule
2c\_vs\_64zg & \multirow{2}{*}{200K} & \colorbox{mine}{\textbf{25.9}\scr{(14.3)}} & 0.5\scr{(0.5)} & 0.0\scr{(0.0)} & 0.1\scr{(0.1)} & 0.0\scr{(0.0)} \\
3s\_vs\_5z & & \colorbox{mine}{\textbf{78.4}\scr{(11.2)}} & 0.0\scr{(0.0)} & 0.0\scr{(0.0)} & 0.0\scr{(0.0)} & 0.0\scr{(0.0)} \\ \midrule
corridor & \multirow{1}{*}{400K} & \colorbox{mine}{\textbf{71.0}\scr{(13.8)}} & 0.0\scr{(0.0)} & 0.0\scr{(0.0)} & 0.0\scr{(0.0)} & 0.0\scr{(0.0)} \\ \bottomrule
\end{tabular}%
}
\label{table:small_t_epsilon}
\end{table}

\subsection{Comparison with existing Transformer-based world models}
\begin{figure*}[t!]
\begin{center}
\centerline{
\includegraphics[width=0.5\columnwidth]{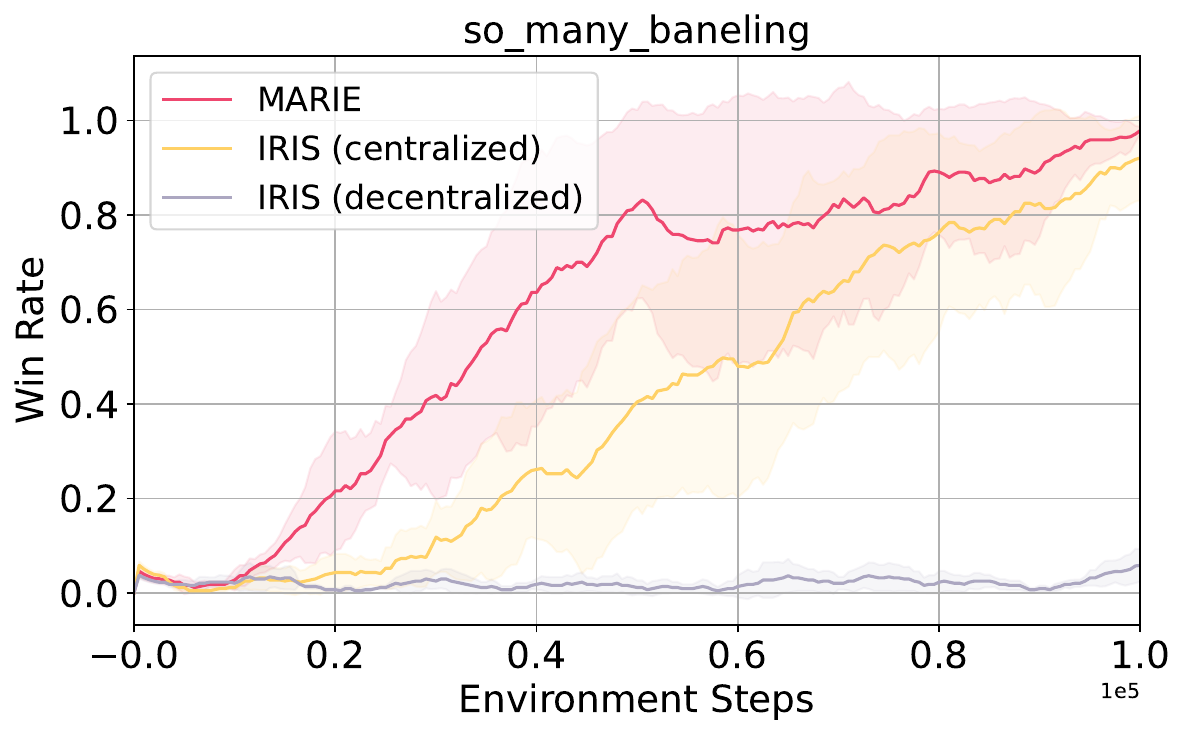}
}
\caption{
{Comparison with two direct extension of IRIS in the \emph{so\_many\_baneling} scenario.}}
\label{fig:iris_comp}
\end{center}
\end{figure*}

{Existing Transformer-based world models are primarily designed for single-agent scenarios, but they can be naturally adapted to multi-agent settings, modeling either independently local dynamics or joint dynamics. Fortunately, we have included IRIS as a Transformer-based world model baseline in our ablation experiments. Specifically, the \emph{Centralized Manner} and \emph{MARIE w/o aggregation} variants from our ablation experiments correspond to IRIS baseline variants under different deployment strategies. But different from their original implementation, these IRIS baseline variants also uses the same actor-critic method as MARIE during learning in imaginations phase (i.e., using PPO instead of REINFORCE for behaviour learning).}
{With a shared behaviour learning phase, we can analyze the reason why existing single-agent Transformer-based world model cannot be directly adapted to MARL. As shown in Figure~\ref{fig:iris_comp}, without incorporating CTDE principle, the learning of single-agent world model would be disrupted by the scalability and non-stationarity issues.}

\subsection{Comparison against MAMBA with Different Imagination Horizon}
{We report the performance of MARIE and MAMBA on \emph{3s\_vs\_5z} with different imagination horizon in Table~\ref{tab:different_H}. And the result shows that the performance gap between the two is not related to the choice of imagination horizon. Interestingly, a larger imagination horizon may help policy learning in imagination. We hypothesize that longer imagined trajectories help alleviating shortsighted behaviours in policy learning.}

\begin{table}[ht!]
\centering
\caption{{The performance of MARIE and MAMBA on \emph{3s\_vs\_5z} with different imagination horizon.}}
\label{tab:different_H}
\vspace{1em}
\begin{tabular}{ccccc}
\toprule
Maps                        & Method & Horizon $H = 8$              & Horizon $H = 15$             & Horizon $H = 25$             \\ \midrule
\multirow{2}{*}{3s\_vs\_5z} & MARIE  & $0.40 \pm 0.34$ & $0.75 \pm 0.09$ & $0.78 \pm 0.11$ \\
                            & MAMBA  & $0.00 \pm 0.00$ & $0.13 \pm 0.14$ & $0.16 \pm 0.13$ \\ \bottomrule
\end{tabular}
\end{table}

\subsection{Comparison between different agent-wise aggregation modules}
To convincingly demonstrate the effectiveness of our designed Perceiver aggregation module, we report the FLOPs for both Perceiver and Self-Attention aggregation with varying number of agents, as shown in Table \ref{table:flops}. Note that the Perceiver aggregation in our algorithm consists of one transformer layer with cross-attention and two transformer layers with self-attention. The details can be found in Table \ref{table:perceiver_hyper}. To make fair comparisons, the Self-Attention aggregation also consists of three transformer layers with self-attention.

\begin{table}[ht!]
\centering
\caption{{The FLOPs of Perceiver aggregation and Self-Attention aggregation with varying number of agents.}}
\label{table:flops}
\vspace{1em}
\begin{tabular}{ccccc}
\toprule
Aggregation Module  & 2 agents & 3 agents & 5 agents & 9 agents \\ \midrule
Perceiver      & 0.016G FLOPs  & 0.024G FLOPs & 0.041G FLOPs & 0.073G FLOPs \\
Self-Attention & 0.133G FLOPs  & 0.201G FLOPs & 0.335G FLOPs & 0.603G FLOPs \\ \bottomrule
\end{tabular}
\end{table}

Additionally, we compare the performance of these aggregation methods in the \emph{2m\_vs\_1z} scenario after 50000 steps. These results show that Perceiver aggregation offers a more computationally efficient solution compared to self-attention aggregation.

\begin{table}[ht!]
\centering
\caption{The performance of different aggregation methods in the \emph{2m\_vs\_1z} scenario after 50000 steps.}
\vspace{1em}
\begin{tabular}{ccc}
\toprule
Map & Perceiver Aggregation (3 layers) & Self-Attention Aggregation (3 layers) \\ \midrule
2m\_vs\_1z & $\mathbf{0.96 \pm 0.07}$  & $0.54 \pm 0.46$ \\ \bottomrule
\end{tabular}
\end{table}

\section{Additional Discussion between CoDreamer and MARIE}\label{appendix:additional_discussion}
{Additionally, a recent method CoDreamer~\citep{toledo2024codreamer} extends DreamerV3~\citep{hafner2023dreamerv3} to the multi-agent setting, using GAT V2~\citep{brody2021attentive} for communication among agents’ world models and policies. Though the aggregation modules in CoDreamer and ours are both built upon the Transformer architecture, our focus lies in computational efficiency of aggregation while it focuses on the underlying topological graph structure among agents.}
{However, a fundamental difference is the backbone used for modeling the local dynamics. While we cast the local dynamics learning as the sequence modeling over discrete tokens, which can be achieved by using auto-regressive Transformers with causal attention mechanism, CoDreamer directly adopts the RSSM framework in DreamerV3.} 

\section{Additional Results with Standardized Performance Evaluation Protocol}\label{appendix:rliable}
\begin{figure*}[h!]
\begin{center}
\centerline{
\includegraphics[width=0.5\columnwidth]{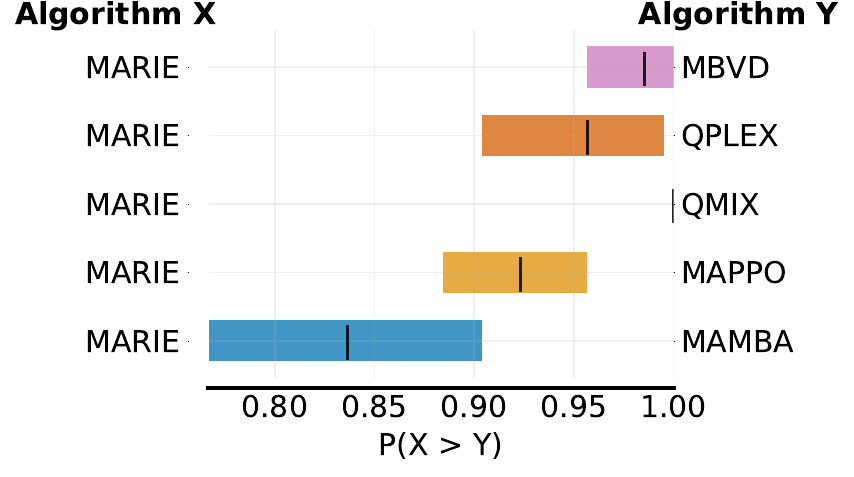}
}
\caption{{Probabilities of improvement} \citep{agarwal2021rliable}.}
\label{fig:prob_improvement}
\end{center}
\end{figure*}

{For finer comparisons, we also provide probabilities of improvement in Figure~\ref{fig:prob_improvement}.}


\newpage
\section{Parameters Setting and Computational Consumption of MARIE}\label{appendix:marie_hyper}
All our experiments are run on a machine with a single NVIDIA RTX 3090 GPU, a 36-core CPU, and 128GB RAM.
We provide the hyperparameters of MARIE for experiments in SMAC, shown as Table~\ref{table:marie_hyper}.
To enable the running of experiments in all SMAC scenarios with a single NVIDIA RTX 3090 GPU, we set the imagination horizon $H$ as 8 for other scenarios involving the number of agents $n > 5$, 15 for $n \leq 5$.
In \emph{so\_many\_baneling} and \emph{2s3z}, we set the imagination horizon $H$ as 5.
Correspondingly, the number of policy updates in imaginations varies with imagination horizon $H$.
As for the scenario \emph{2c\_vs\_64zg}, considering the significantly large action space in it, we enable the observation of agent id and last action for each agent and disable stacking the last 5 observations as input to the policy.

Based on the above reported setting, we present a rough computational consumption in Table~\ref{table:overhead_analysis}.

\begin{table}
  \caption{Hyperparameters for MARIE in SMAC environments.}
  \label{table:marie_hyper}
  \centering
  \begin{tabular}{ll}
    \toprule
    {\bf Hyperparameter}   &  {\bf Value}                   \\
    \midrule
    Batch size for tokenizer training     &    256       \\
    Batch size for world model training     &    30       \\
    Optimizer for tokenizer         &   AdamW           \\
    Optimizer for world model       &   AdamW           \\
    Optimizer for actor \& critic   &   Adam            \\
    Tokenizer learning rate         &   0.0003          \\
    World model learning rate       &   0.0001          \\
    Actor learning rate             &   0.0005          \\
    Critic learning rate            &   0.0005          \\
    Gradient clipping for actor \& critic          &     100             \\
    Gradient clipping for tokenizer          &     10             \\
    Gradient clipping for world model        &     10             \\
    Weight decay for world model    &   0.01            \\
    $\lambda$ for $\lambda$-return computation     &    0.95      \\
    Discount factor $\gamma$        &   0.99            \\
    Entropy coefficient             &   0.001           \\
    Buffer size (transitions)       &   $2.5\times 10^5$          \\
    Number of tokenizer training epochs          &     200             \\
    Number of world model training epochs        &     200             \\
    Collected \underline{transitions} between updates        &     \{100, 200\}    \\
    Epochs per policy update (PPO epochs)        &     5               \\
    PPO Clipping parameter $\epsilon$            &     0.2             \\
    Number of imagined rollouts     &     600 or 400    \\
    Imagination horizon $H$         &     \{15, 8, 5\}  \\
    Number of policy updates        &     \{4, 10, 30\} \\ \midrule
    Number of stacking observations &     5             \\
    Observe agent id                &     False         \\
    Observe last action of itself   &     False         \\
    \bottomrule
  \end{tabular}
\end{table}

\begin{table}
  \caption{Computational time consumption of MARIE in SMAC.}
  \centering
  \begin{tabular}{cccc}
    \toprule
    Environment Steps & 100000 & 200000 & 400000   \\ \midrule
    Training Time     &  1 day & 2-3 days & 4 days \\
    Usage of GPU Mem  &  22GB  & 22GB   & 22GB         \\
    \bottomrule
  \end{tabular}
  \label{table:overhead_analysis}
\end{table}

\clearpage
\section{Overview of MARIE Algorithm}\label{appendix:algo}
Pseudo-code is summarized as Algorithm~\ref{alg:marie}.
\begin{algorithm} 
\caption{MARIE}
\label{alg:marie}
\begin{algorithmic}[0]
    \STATE \textcolor{gray}{// main loop of training} 
    \FOR{\textit{epochs}}
        \STATE collect\_experience(\textit{num\_transitions})
        \FOR{\textit{learning\_world\_model\_steps\_per\_epoch}}
            \STATE train\_world\_model()
        \ENDFOR
        \FOR{\textit{learning\_behaviour\_steps\_per\_epoch}}
            \STATE train\_agents()
        \ENDFOR
    \ENDFOR
    \STATE 
    \STATE {\bf function} collect\_experience($n$):
    \STATE $\boldsymbol{o}_{0} \leftarrow$ env.reset()
    \FOR{$t = 0, \ldots, n - 1$}
        \STATE \textcolor{gray}{// processed by VQ-VAE}
        \STATE $\boldsymbol{\hat{o}}_{t} \leftarrow D(E(\boldsymbol{o}_{t}))$
        \STATE Sample $a_{t}^{i} \sim \pi_{\psi}^{i}(a_{t}^{i} | \hat{o}_{t}^{i}) \,, \forall i$
        \STATE $\boldsymbol{o}_{t + 1}, r_t, done \leftarrow$ env.step($\boldsymbol{a}_{t}$)
        \IF{$done = True$}
            \STATE $\boldsymbol{o}_{t+1} \leftarrow$ env.reset()
            \STATE $\gamma_t \leftarrow 0.$
        \ELSE
            \STATE $\gamma_t \leftarrow 0.99$
        \ENDIF
    \ENDFOR
    \STATE $\mathcal{D} \leftarrow \mathcal{D} \cup \{ \boldsymbol{o}_{t}, \boldsymbol{a}_{t}, r_{t}, \gamma_t \}^{n - 1}_{t = 0}$
    \STATE
    \STATE {\bf function} train\_world\_model():
    \STATE Sample $\{ \boldsymbol{o}_{t}, \boldsymbol{a}_{t}, r_{t}, \gamma_t \}^{t = \tau + H - 1}_{t = \tau}$
    \STATE Update $(E, D, \mathcal{Z})$ via $\mathcal{L}_{\rm VQ-VAE}$ over observations $\{ \boldsymbol{o}_{t} \}^{t = \tau + H - 1}_{t = \tau}$
    \FOR{agent $i = 1, \ldots, n$}
        \STATE Update $\phi, \theta$ via $\mathcal{L}_{\rm Dyn}(\phi, \theta)$ over local trajectories $\{ {o}_{t}^{i}, {a}_{t}^{i}, r_{t}, \gamma_t \}^{t = \tau + H - 1}_{t = \tau}$
    \ENDFOR
    \STATE
    \STATE {\bf function} train\_agents():
    \STATE Sample an initial observation $\boldsymbol{o}_{0} \sim \mathcal{D}$
    \STATE $\{x_{0, j}^{i}\}_{j = 1}^{K} \leftarrow E(o_{0}^{i}), \hat{o}_{0}^{i} \leftarrow D(E(o_{0}^{i})) \,, \forall i$
    \FOR{$t = 0, \ldots, H - 1$}
        \STATE Sample $a_{t}^{i} \sim \pi_{\psi}^{i}(a_{t}^{i} | \hat{o}_{t}^{i})\,,\forall i$
        \STATE Aggregate $( x_{t, 1}^{1}, \ldots, x_{t, K}^{1}, a^{1}_{t}, \ldots, x_{t, 1}^{n}, \ldots, x_{t, K}^{n}, a^{n}_{t} )$ into $(e_{t}^{1}, \ldots, e_{t}^{n})$ via the Perceiver $\theta$
        \STATE Sample $\hat{x}_{t + 1, \cdot}^{i}, \hat{r}_{t}^{i}, \hat{\gamma}_{t}^{i} \sim p_{\phi} (\hat{x}_{t + 1, \cdot}^{i}, \hat{r}_{t}^{i}, \hat{\gamma}_{t}^{i} | x_{0, \cdot}^{i}, a_{0}^{i}, e_{0}^{i}, \ldots, \hat{x}_{t, \cdot}^{i}, a_{t}^{i}, \hat{e}_{t}^{i})\,,\forall i$
        \STATE $\hat{o}_{t + 1}^{i} \leftarrow D(\hat{x}_{t + 1, \cdot}^{i})\,,\forall i$
    \ENDFOR
    \FOR{agent $i = 1, \ldots, n$}
        \STATE Update actor $\pi_{\psi}^{i}$ and critic $V_{\xi}^{i}$ via $\mathcal{L}_{\rm Dyn}(\phi, \theta)$ over imagined trajectories $\{ \hat{o}_{t}^{i}, {a}_{t}^{i}, \hat{r}_{t}^{i}, \hat{\gamma}_{t}^{i} \}^{t = H - 1}_{t = 0}$
    \ENDFOR
\end{algorithmic}
\end{algorithm}



\end{document}